\documentclass[preprint,12pt]{elsarticle}
\usepackage{amsmath,amssymb,amsfonts}
\usepackage{algorithmic}
\usepackage{amssymb}
\usepackage{graphicx}
\usepackage{xcolor}
\usepackage{textcomp}
\usepackage[colorlinks]{hyperref}
\usepackage{changepage}
\usepackage[colorlinks]{hyperref}
\usepackage{soul}
\usepackage{booktabs}
\usepackage{csquotes}
\usepackage{float}
\begin{document}
\begin{frontmatter}
\title{AGA-GAN: Attribute Guided Attention Generative Adversarial Network with U-Net for Face Hallucination}


\author[inst1]{Abhishek Srivastava}
\ead{abhisheksrivastava2397@gmail.com}

\affiliation[inst1]{organization={Computer Vision and Pattern Recognition Unit,Indian Statistical Institute},%Department and Organization
            city={Kolkata},
            postcode={700108},
            state={West Bengal},
            country={India}}

\author[inst2]{Sukalpa Chanda}
\ead{sukalpa@ieee.org}
\author[inst1]{Umapada Pal}
\ead{umapada@isical.ac.in}

\affiliation[inst2]{organization={Department of Computer Science and Communication, Østfold University College},%Department and Organization
            city={Halden},
            country={Norway}}

\begin{abstract}
The performance of facial super-resolution methods relies on their ability to recover facial structures and salient features effectively. Even though the convolutional neural network and generative adversarial network-based methods deliver impressive performances on face hallucination tasks, the ability to use attributes associated with the low-resolution images to improve performance is unsatisfactory.
In this paper, we propose an Attribute Guided Attention Generative Adversarial Network which employs novel attribute guided attention (AGA) modules to identify and focus the generation process on various facial features in the image. Stacking multiple AGA modules enables the recovery of both high and low-level facial structures. We design the discriminator to learn discriminative features exploiting the relationship between the high-resolution image and their corresponding facial attribute annotations. We then explore the use of U-Net based architecture to refine existing predictions and synthesize further facial details. Extensive experiments across several metrics show that our AGA-GAN and AGA-GAN+U-Net framework outperforms several other cutting-edge face hallucination state-of-the-art methods. We also demonstrate the viability of our method when every attribute descriptor is not known and thus, establishing its application in real-world scenarios.
\end{abstract}
\end{frontmatter}

%%       \citet{<label>} ==> Jones et al. [21]
%%       \citep{<label>} ==> [21]
%%

\section{Introduction}
%We address the challenge of Face Hallucination which is reconstructing a High resolution facial image from an affiliated Low resolution counter part.
Face Hallucination is a domain-specific super-resolution task that aims to learn the mapping between low resolution (LR) image and its corresponding High resolution (HR) counterpart. Its popularity lies in the fact that it has great utility in a variety of applications like surveillance, face recognition, and expression recognition. Contrary to generic image super-resolution, the performance of face hallucination models relies heavily on its ability to effectively recover facial details and attributes. Maintaining the structural integrity of the face is imperative as the absence of it can lead to unnatural artifacts which may hamper the quality of the image. Particularly in surveillance, it has been empirically proven~\cite{zou2011very} that a minimum resolution ranging from 32 x 32 to 64 x 64 serves as a prerequisite for effective face recognition systems. In reality, video surveillance equipment may not be able to capture facial images satisfying the above criteria. This warrants a face hallucination architecture that can effectively upsample LR images. 
%\begin{figure}[!t]
%    \centering
%    \includegraphics[width=0.5\textwidth]{figures/framework.jpeg}
%    \caption{Workflow of our AGA-GAN and U-Net framework}
%    \label{fig:framework}
%\end{figure}
In recent times this problem has drawn great attention ~\cite{ma2010hallucinating}. It is worth mentioning here that while performing an 8x upsampling operation to produce an HR image, a method has only access to its LR counterpart which contains 1.56\% pixels of the original image. While interpolation-based face hallucination techniques~\cite{zhang2008image,sun2010gradient} lacked the spatial details and visual quality, the presence of a large amount of HR and corresponding LR images enabled learning-based techniques~\cite{zhang2015learning,chang2004super} to deliver promising results. The application of CNN~\cite{dong2016accelerating,shi2016real} and GAN~\cite{yu2016ultra,hsu2019sigan} based methods demonstrated superior performance with high visual quality and accurate structural representations. While usual methods exploit the prior structural information of the LR images in their generation process, the presence of attribute descriptors or labeled feature annotations can be further leveraged to enhance the visual quality of super-resolved images.
In this paper. we propose AGA-GAN(Attribute Guided Attention Generative Adversarial Network) which uses a novel attention mechanism to identify the spatial location of key facial attributes and focus the generation process to successfully recover salient facial elements. The AGA-GAN allows the attribute descriptors to learn their spatial position in the image domain using the attribute stream. The attribute streams interact with the main generation stream to allow feature maps carrying information about several facial elements to be incorporated in the final HR prediction. Attention maps generated by the attribute stream enable the main generation stream to focus on the spatial locations of essential features and also provide additional information about the description of facial elements. Feedback from the main generation stream allows the attribute stream to generate attention maps progressively focusing on various facial structures. Consequently, the super-resolved (SR) prediction possesses a high degree of fidelity and enjoys high-frequency details demonstrating superior visual quality. Apart from that, we design a spatial and channel attention-based U-Net~\cite{ronneberger2015u} for enhancing the visual quality by refinement of facial features and even rectification of visual artifacts present in SR prediction by AGA-GAN (check Figure~\ref{fig:U-Net_q}). It can be noted that apart from demonstrating a high degree of fidelity, the use of the U-Net module successively after AGA-GAN improves the quantitative performance (see Section~\ref{section:refinementU-Net}).

The contributions of our paper are summarized below:
%We propose AGA-GAN(Attribute Guided Attention Generative Adversarial Network) which uses a novel attention-based mechanism that uses the attributes to generate attention maps to progressively refine the facial features. This allows the generation process of the HR image to focus on the structure and details of the facial features. We further propose a spatial and channel attention U-Net that refines the generated SR image. The contributions of our paper are summarized below:
\begin{enumerate}
    \item We propose an attribute-guided attention-based technique that progressively refines the facial features in higher spatial dimensions to improve the quality of the generated SR image.
    \item We propose a discriminator which leverages prior attribute descriptors to learns the mapping between attributes and the HR image to further increase its capability to determine whether the image is real or generated.
    \item We show the superiority of AGA-GAN in real-world scenarios where only partial attribute descriptors will be present.
    \item We propose a U-Net-based architecture to take the prediction of our AGA-GAN and refine the image further as well as to increase the perceptual quality of the image.
    \item Exhaustive experiments demonstrate that AGA-GAN and AGA-GAN+U-Net outperform the previous state-of-the-art methods on all standard metrics.
\end{enumerate}
The organization of the rest of the paper is as follows. Section~\ref{section:realatedwork} presents a brief literature survey of methods developed for face hallucination. Section~\ref{section:method} describes our proposed AGA-GAN+UNet framework. Our entire experiment setup in described in Section~\ref{section:experiments} and the results are reported in Section~\ref{section:results}. Finally the conclusion is presented in Section~\ref{section:conclusion}.

\section{Related Work}
\label{section:realatedwork}
\subsection{Face Hallucination}
Face hallucination has been explored over the past few decades and was originated by Baker and Kanade~\cite{baker2000hallucinating}, where a multi-level learning model based on Gaussian image pyramid was used to increase the resolution of LR images. Liu et al.~\cite{liu2001two} used principal component analysis(PCA) and Markov random field(MRF) for face hallucination. Ensuring these pioneering works, various global face statistical methods~\cite{gunturk2003eigenface,wang2005hallucinating,liang2013face} and local patch-based methods~\cite{chang2004super,ma2010hallucinating} have been introduced for super-resolution of faces. Zhou et al.~\cite{zhou2015learning} initially used CNN to propose a bi-channel convolutional neural network (BCCNN) to hallucinate global face images. Chen et al.~\cite{chen2020rbpnet} proposed Residual back-projection network
(RBPNet) which used a base model to extract features for face hallucination and edge map prediction boundaries. Huang et al~\cite{huang2017wavelet} exploited the fact that wavelet transform potentially represents the contextual and textural information of the image to design WaveletSRNet.They transformed the LR face images to wave coefficients and super-resolved the face image in the wavelet coefficient domain. Yu et al.~\cite{yu2016ultra} leverage GAN~\cite{goodfellow2014generative} to develop URDGN which increased the perceptual quality of HR image. Indradi et al.~\cite{indradi2019face} used inception residual networks inside the GAN framework to boost performance and stabilize training. HiFaceGAN~\cite{yang2020hifacegan} uses a suppression module for the selection of informative features which are then used by a replenishment module for recovery detail. SiGAN~\cite{hsu2019sigan} uses two identical generators with pair-wise contrastive loss based on the fact that different LR face images possess different identities. Hence, they were able to super-resolved LR facial images while preserving their identities. Jiang et al.~\cite{jiang2019atmfn} used different deep learning-based approaches such as CNN, RNN, GAN to generate candidate SR images. Rather than using a pre-determined technique to combine the candidate predictions, ATMFN used an attention sub-network to learn the individual fusion weight matrices to determine useful components of candidate SR images. Further, a threshold-based fusion and reconstruction module combines the candidate HR image to give the final SR prediction. Chen et al.~\cite{chen2020learning} proposed facial attention units(FAUs) which used a spatial attention mechanism to learn and focus on different face structures.
\subsection{Attention Networks}
Attention mechanisms are useful in identifying the most relevant features necessary for the effective completion of a task. Recently, this mechanism has been of great interest in the field of computer vision and has been studied extensively. Mnih et al.~\cite{mnih2014recurrent} devised a model capable of identifying a sequence of regions that convey the most relevant information. Hu et al.~\cite{hu2018squeeze} proposed SE-Net which pioneered channel-wise attention. The Squeeze and Excitation ($\text{S\&E}$) block was able to model interdependencies between the channels and derive a global information map that helps in emphasizing relevant features and suppressing irrelevant features. Xu et al.~\cite{xu2015show} devised a dual attention-based module for image captioning. Wang et al.~\cite{wang2017residual} proposed to generate attention-aware features by stacking attention modules. Woo et al.m~\cite{woo2018cbam} generated spatial and channel attention feature maps to multiply with input feature maps and refine them. Fu et al.~\cite{fu2017look} used an attention module that uses previous predictions as a reference while sequentially generating region attention.
In the field of face hallucination, ATMFN~\cite{jiang2019atmfn} used an attention module to recognize the most relevant regions from all candidate SR images. SPARNet~\cite{chen2020learning} used spatial attention in the generation process to progressively focus on various facial structures.

\section{Methodology}
\label{section:method}
\begin{figure*}[!t]
    \centering
    \includegraphics[width=1\textwidth]{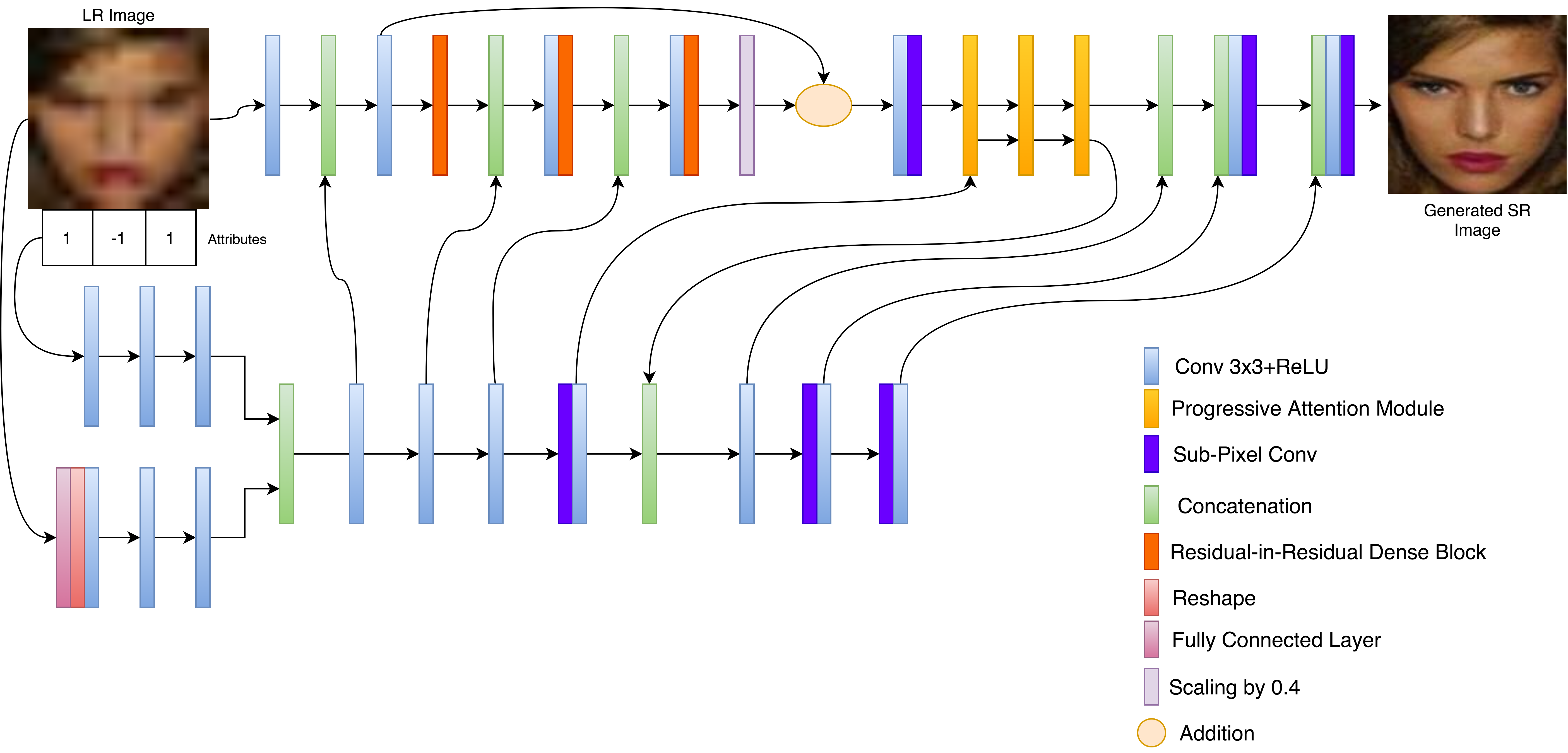}
    \caption{The Attribute Guided Attention Generator  Architecture}
    \label{fig:generator}
\end{figure*}

In this section, we present our proposed attribute-guided attention generative adversarial network. Here, LR, HR, SR, att, and s represent the low-resolution input, high-resolution target image, super-resolved prediction, attribute descriptors, and the upscale factor respectively. We explain various components of our two-stream generator. The generator aims to upscale the prior low resolution image LR (where $LR\, \in \, R^{W/s\:\times\:H/s\:\times\:3}$) to high resolution space SR (where $SR\, \in \, R^{W\:\times\:H\:\times\:3}$) with the assistance of att (where $att$ $\in \, R^{38}$). 
The generator comprises a mainstream and an attribute stream (Figure~\ref{fig:generator}). The mainstream takes the LR as input while the attribute stream takes the pair (LR and att.) as input. The attribute stream subsequently helps the main stream in generating an accurate high-resolution image with high fidelity. The attribute-guided attention module generates attention maps that guide the main stream to focus on the regions of facial attributes and maintains the structural integrity of the image. The feedback from the main stream to the attribute stream allows the attribute-guided attention maps to progressively create attention maps on various facial elements. Stacking these modules allows an increased receptive field of the attention maps enabling the generation stream to focus on both high and low-level facial structures. The residual structure of the module leads to faster convergence and improved information flow. We also propose a discriminator which uses att as a prior and tries to distinguish between all (HR, att) and (SR, att) pairs. The super-resolved image from the AGA-GAN along with bicubic interpolation of the LR image is concatenated and is then fed into spatial and channel attention U-Net. The U-Net aims to refine existing facial structures and add further high-frequency details to raise the perceptual quality and richness of generated SR predictions. We expand on our proposed approach in the subsequent sections.

\subsection{Generator}
The generator consists of two streams denoted as the main stream and attribute stream (see Figure~\ref{fig:generator}). The main stream takes an LR image as input and the attribute stream takes an attribute vector as well as the corresponding LR image. The attribute vector provides information about the presence and absence of various facial features in the image. The main stream initially has 3 convolutional and ReLU layers. We then use residual dense blocks proposed in ~\cite{zhang2018residual} to increase the effective receptive fields and extract high-level features from the input image while preserving low-level features. The extracted features are scaled by factor and added back to the input to avoid instability~\cite{szegedy2017inception,lim2017enhanced}. The residual design of the blocks allows relevant high and low-level features to be preserved for effective up-sampling. The attribute stream translates the one-dimensional vector to two-dimensional image space and forces interaction with LR image stream embedded in the attribute stream (Figure~\ref{fig:generator}), thus enabling the features to capture the relationship between the facial attribute descriptors and the image. We then use sub-pixel convolution~\cite{shi2016real} to increase the spatial dimension by a factor of 2 simultaneously in the main and the attribute stream.  Further, the attribute-guided attention module uses the attribute stream's feature maps to generate attention coefficients to learn the spatial correlation between attributes and their corresponding location in the image domain. This enables the main stream to improve the facial and textural features of the SR image.
\subsubsection{Attribute Guided Attention Module}
\begin{figure*}[!t]
    \centering
    \includegraphics[width=1\textwidth]{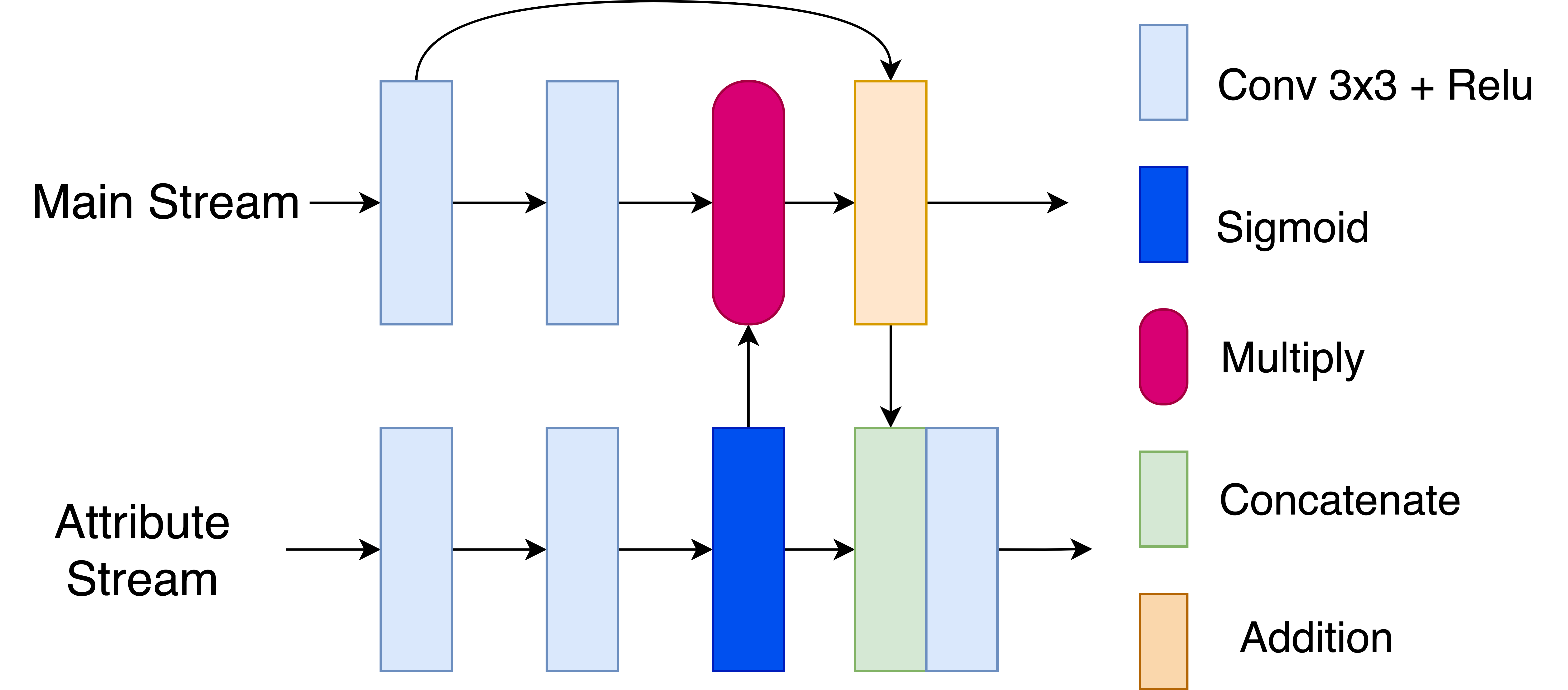}
    \caption{The Attribute Guided Attention Module}
    \label{fig:agatt}
\end{figure*}

In this section, we describe our attribute guided attention module (see Figure~\ref{fig:agatt}). The attribute stream utilizes the attribute vector and the LR image to learn the mapping between the features and their region of interest in the image domain. Initially, two stacked convolutional layers and a single convolutional layer are used to operate on sets of feature maps extracted from the main and attribute stream respectively, as explained in Equation~\ref{eq:pat11} and Equation~\ref{eq:pat12}
\begin{equation}{\label{eq:pat11}}
SR_{main} = Conv(Conv(SR_{main}))
\end{equation}
\begin{equation}{\label{eq:pat12}}
AS_{stem} = Conv(AS_{stem})
\end{equation}
We use the sigmoid activation function to calculate an attention coefficient for each spatial location in the feature maps as described in Equation~\ref{eq:pat21}. 
\begin{equation}{\label{eq:pat21}}
AGA = \sigma(Conv_{1x1}(AS_{stem}))
\end{equation}
The $AGA$ map is multiplied with $SR_{main}$ to identify the regions in the image domain corresponding to the facial attributes. This is termed as attribute enhanced feature maps in Equation~\ref{eq:pat22} which is then added back to the $SR_{main}$ (Equation~\ref{eq:pat23}). This residual connection preserves the relevant low-level structural information and improves convergence.
\begin{equation}{\label{eq:pat22}}
AEF = AGA \otimes SR_{main}
\end{equation}
\begin{equation}{\label{eq:pat23}}
SR_{main} =SR_{main} + AEF
\end{equation}
The refined $SR_{main}$ is then fused with $AS_{stem}$ to provide feedback to the attribute stream in Equation~\ref{eq:pat3}. This allows subsequent attribute-guided attention modules to progressively generate attention maps focusing on auxiliary facial elements. Consecutive attribute guided attention modules also provide an additional advantage in identifying high-level target structures by increasing the receptive field. We use three such modules in a sequence.
\begin{equation}{\label{eq:pat3}}
AS_{stem} = AS_{stem} \oplus SR_{main}
\end{equation}

\subsubsection{De-convolution Sub-Network}
The resultant feature maps by attribute guided attention modules of both main stream ($SR_{main}$) and attribute stream($AS_{stem}$) are used for upscaling. The $SR_{main}$ and $AS_{stem}$ are both upsampled using sub-pixel convolution~\cite{shi2016real}. The main stream feature maps are then concatenated with attribute stream feature maps as described in Equation~\ref{eq:pat4}, this helps to further leverage the attribute information propagated in the upsampling process (refer Figure~\ref{fig:generator}). We upsample the feature maps twice in the case of 8x upsampling and once in the case of 4x upsampling.

\begin{equation}{\label{eq:pat4}}
    SR_{main} = SR_{main} \oplus AS_{stem}
\end{equation}

\begin{figure*}[!t]
    \centering
    \includegraphics[width=1\textwidth]{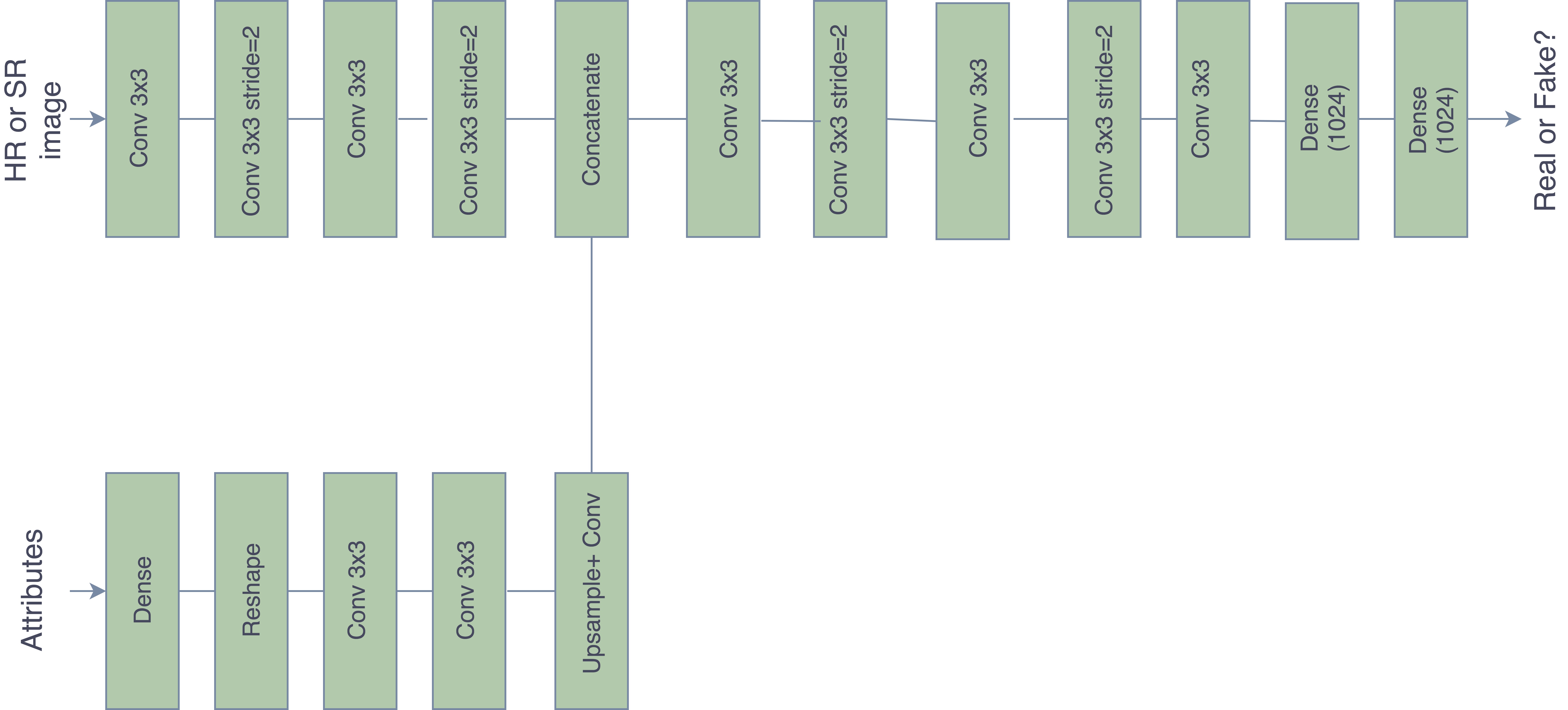}
    \caption{Discriminator Architecture}
    \label{fig:discriminator}
\end{figure*}
\subsection{Discriminator}
We design our discriminator to take both the HR/SR image and the attribute descriptors of the corresponding image as input. This enables the discriminator to learn the correlation between the facial features and structures of the HR image and its corresponding attribute descriptors.
The attribute prior then enables the discriminator to distinguish between the real HR image and the fake SR image based on that facial feature's existence in the image domain as well as its shape and composition. There are again two branches that process the image and the attribute vector separately (see Figure~\ref{fig:discriminator}). The attribute vector is first connected to a fully connected layer and is then reshaped into 16x16x3. The HR/SR image is fed into the main branch which comprises convolutional units. Each unit consists of a convolutional layer with a 3x3 kernel with LeakyReLU activation. This is followed by a convolutional layer with a kernel size of 3 and stride of 2 and a LeakyReLU activation function. When the image spatial dimensions are reduced from 128x128 to 32x32, the tensors from the main branch and attribute branch are concatenated to combine the information from the image domain and its attribute descriptors. This enables the relationship between the attribute priors and their presence, structure, and composition in the corresponding image to serve as discriminative features in distinguishing the real and the generated image.
\begin{figure*}[!t]
    \centering
    \includegraphics[width=1\textwidth]{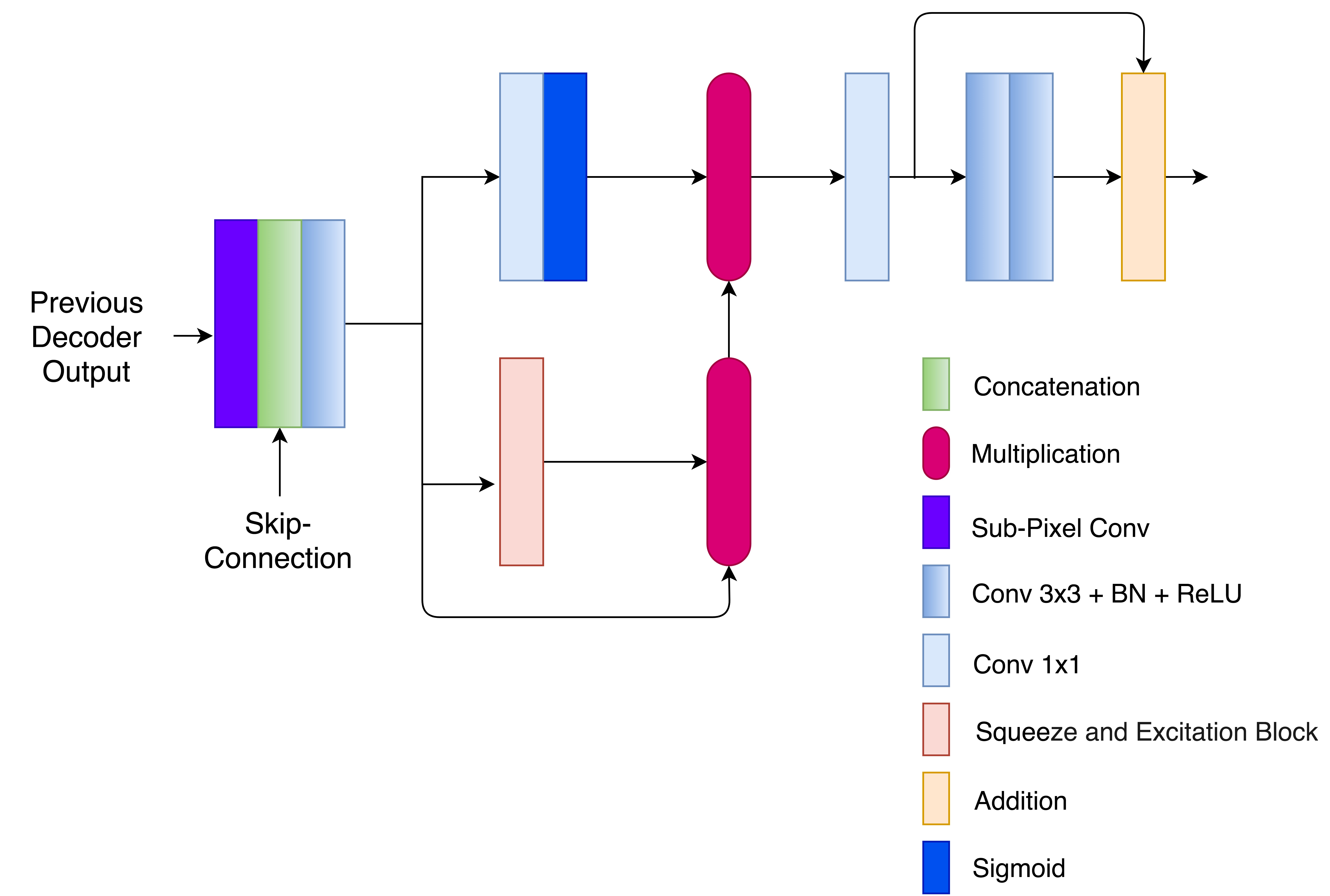}
    \caption{Structure of Dual Attention Block}
    \label{fig:dual}
\end{figure*}
\begin{figure*}[!t]
    \centering
    \includegraphics[width=1\textwidth]{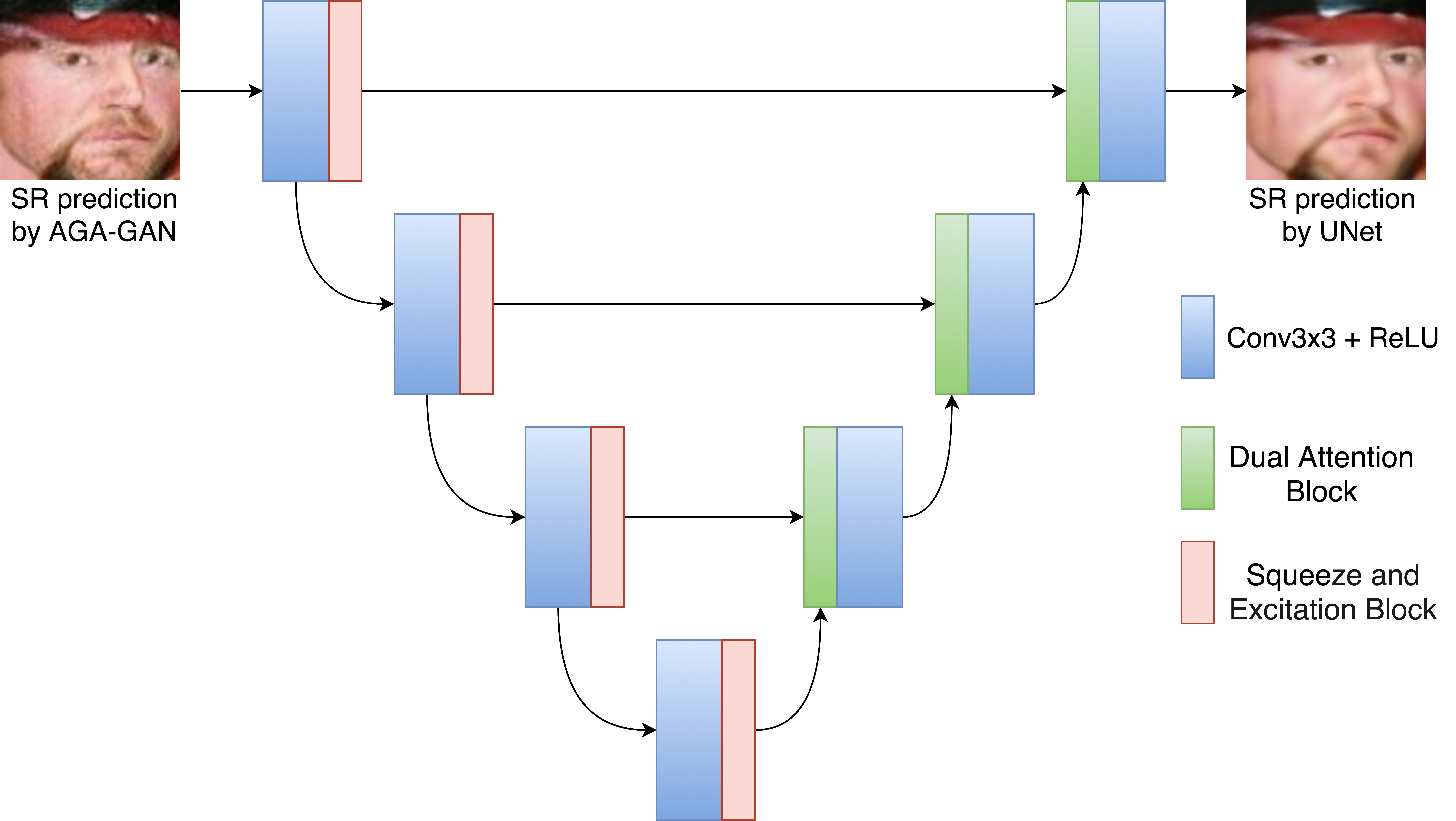}
    \caption{Dual attention U-Net Architecture}
    \label{fig:U-Net}
\end{figure*}
\subsection{U-Net}
We build a U-Net architecture (see Figure~\ref{fig:U-Net}) which takes the SR prediction of AGA-GAN and bicubic interpolation of the LR image as input and uses the channel and spatial attention in the encoder-decoder architecture. The bicubic interpolation of the LR image is combined with the SR image by concatenation along the channel axis. We use an encoder-decoder-based network to refine the pre-existing SR image. The encoder block consists of convolutional layers followed by LeakyReLU activation and ($\text{S\&E}$) layer. The  ($\text{S\&E}$) block in the network increases the network's representative power by computing the interdependencies between channels. During the squeezing step, global average pooling is used to aggregate feature maps across the channel's spatial dimensions. In the excitation step, a collection of per-channel weights are produced to capture channel-wise dependencies~\cite{hu2018squeeze}. At each encoder stage, max pooling with the stride of 2 is used for downscaling the resolution, and drop out is used for the model regularization. The skip connections propagate low-level features from the encoder block to its corresponding decoder block from the same level. We also use dual attention blocks (see Figure~\ref{fig:dual}) in the decoder level to provide spatial and channel attention to the features transmitted by the previous decoder blocks. The first attention mechanism applies channel attention, whereas the second attention uses a spatial attention mechanism. We have used a  $\text{S\&E}$ block for the calculation of channel-wise scale coefficients denoted by $X_{se}$. Spatial attention is also calculated at the same top stream where the input channels $\mathit{C}$ are reduced to $1$ using $1\times1$ convolution. The sigmoid activation functions $\sigma(\cdot)$ is used to scale the values between 0 and 1 to produce an activation map, which is stacked $C$ times to give $X_{c}$. The output of the spatial and channel attention can be represented as:
\begin{equation}{\label{eq:decodersc}}
D_{sc} = (X_{c}+1)\otimes  X_{se},
\end{equation}
\noindent{where} $\otimes$ denotes the Hadamard product and $X_{c}$ is increased by a magnitude of 1 to amplify relevant features determined by the activation map. $D_{sc}$ represents the output of the dual attention block which is then fed into the decoder block which consists of convolutional layers of 3x3 kernel size followed by LeakyReLU activation and sub-pixel convolutional layers to increase the spatial dimensions by a factor of 2. The output of the final decoder block is processed by a 1x1 convolutional layer which decreases the output number of channels to 3 and applies tanh activation. We find that this improves the performance of the network even further and increases the fidelity and visual quality of the generated images (see Section~\ref{section:refinementU-Net}).
\subsection{Loss Function}
We denote the SR image prediction by AGA-GAN as $SR_{AGA}$ and the SR prediction by U-Net as $SR_{U}$. The initial discriminator output is $D(x)$ which is the probability that the input image $x$ is real. The loss for the discriminator is given as:
\begin{equation}{\label{eq:dloss}}
L_{D} =  \sum_{n=1}^{N} [-\log(D(HR) - \log(1-D(SR_{AGA})]
\end{equation}
The loss for the generator is given as:
\begin{equation}{\label{eq:gloss}}
L_{G} =\sum_{n=1}^{N} \log(1-D(SR_{AGA})
\end{equation}

where, N represents the total data samples.
Pixel-wise mean absolute error (MAE) loss is calculated as:
\begin{equation}{\label{eq:mse}}
L_{mae}(P,Q) = \frac{1}{W\times H}\sum_{x=1}^{W}\sum_{y=1}^{H} \lvert P_{x,y} - Q_{x,y}\rvert
\end{equation}
We use VGG-19~\cite{simonyan2014very} pre-trained on ImageNet~\cite{deng2009imagenet} to gather feature representations of the input image. The Euclidean distance between the feature representations of the HR image and the SR image serves as the perceptual loss $L_{percep}$.
%The features extracted by feeding the HR image is treated as the ground truth forcing the generator to produce images mimicking the same perceptual quality and hence, enhancing the visual quality.
The total loss function of the generator is defined as:
\begin{equation}{\label{eq:AGAloss}}
\begin{split}
L_{total} = 0.003*L_{GAN} + L_{percep}(HR,SR_{AGA}) \\ 
+ L_{mae}(HR,SR_{AGA}) &
\end{split}
\end{equation}
Similarly the loss in the U-Net stage is,
\begin{equation}{\label{eq:Uloss}}
L_{U-Net} =L_{percep}(HR,SR_{U}) + L_{mse}(HR,SR_{U}) 
\end{equation}
\section{Experiments}
\label{section:experiments}
\subsection{Implementation Details}
We use one-sided label smoothing~\cite{salimans2016improved} to encourage the discriminator to predict soft probabilities. Preventing the discriminator from being overconfident in its prediction stabilizes the training and also provides regularization~\cite{szegedy2016rethinking}. 
MTCNN~\cite{zhang2016joint} is used for face alignment as the initial pre-processing step. We resize the images to 128 x 128 and downsample them to 16 x 16 and 32 x 32 for 8x and 4x upscaling tasks respectively, before feeding it into the model where the batch size used is 50. The images are normalized to a range of [-1,1] and tanh activation is used to generate the final prediction which maps the image in the range of [-1,1]. The output of residual in residual dense blocks is scaled by a factor of 0.4 before adding it into the input. The Adam optimizer was used with a constant learning rate of $1e^{-4}$, and a dropout regularization with $p=0.2$ was used. All models are trained for 50 epochs.
\subsection{Evaluation metrics}
Standard computer vision metrics for Face hallucination such as Peak signal-to-noise ratio (PSNR), Structural Similarity Index (SSIM) have been used for our experiments. We have also used Feature-based similarity index (FSIM)~\cite{zhang2011fsim}, Signal to reconstruction error ratio (SRE)~\cite{lanaras2018super}, Universal image quality index (UIQ)~\cite{wang2002universal} for a detailed and thorough comparison with other SOTA methods. Additionally, we have provided BRISQUE score for all methods, which is a no-reference metric for evaluating image quality. The BRISQUE score of an image is inversely proportional to the image quality~\cite{mittal2012no}.

\subsection{Dataset}
We use CelebA dataset~\cite{liu2015faceattributes} for our experiments.
The CelebA dataset consists of 202,599 images and each image is accompanied by 40 binary attribute annotations. The images carry large fluctuations in pose and background clutter. For our experiments, we choose 100,000 images for training and 10,000 images for testing. Each image has 38 attributes that describe the features of the image as 2 classes fall out of the region after preprocessing. For attribute-specific evaluation, 1000 images that have \enquote{big nose}, 500 images with \enquote{eyeglasses}, 400 images that possess a \enquote{goatee}, 500 images with a \enquote{mustache}, and 1000 images that have \enquote{narrow eyes} are selected for the experiments.
\section{Results and Discussion}
\label{section:results}
To demonstrate the effectiveness of our proposed method we have conducted several experiments which are structured as follows. We report the performance of our method in 8x upscaling and 4x upscaling tasks in Sub-section~\ref{section:upscale8} and Sub-section~\ref{section:upscale4}, respectively to exhibit the superior performance of AGA-GAN and AGA-GAN+U-Net in face hallucination tasks in comparison to other published methods.
We also evaluate the performance of our method on attribute-specific images in Sub-Section~\ref{section:local_region} to illustrate the network's capacity to recover challenging facial attributes. We compare the advantages of using U-Net with AGA-GAN for face hallucination in Sub-section~\ref{section:refinementU-Net}. In Sub-section~\ref{section:scarceatt}, we analyze the performance of our proposed method when only partial attributes are present.

To show further practicality of our method in real-world scenarios, we study the performance of AGA-GAN when partial attributes are known in Sub-section~\ref{section:scarceatt}.
\subsection{Comparison with Other Published Methods}
In this section, we describe the methods that are used for the quantitative and qualitative comparison with AGA-GAN and AGA-GAN+U-Net.
\begin{itemize}
    \item AACNN~\cite{lee2018attribute} also uses the attributes and their interaction with LR image for the generation of SR predictions.
    \item SPARNet~\cite{chen2020learning}, which uses a spatial attention mechanism to focus the generation process on key face structure regions.
    \item ATMFN~\cite{jiang2019atmfn} which unified CNN, RNN, and GAN-based super-resolvers and an attention sub network to gather the most informative regions. The combination of these candidates was formed using threshold-based fusion to give the final HR prediction.
    \item SiGAN~\cite{hsu2019sigan} uses dual generators and discriminators which ensures the preservation of identities while making the HR prediction.
\end{itemize}
We use the author-released codes of these methods for our experiments. Further, these models are trained under the same settings and follow the same train-test split to provide a fair comparison. In all the Tables reporting the quantitative comparison of our method with other state-of-the-art methods, \textbf{The best performance is in bold and second-best performance is highlighted in blue}
\subsubsection{Comparison with the scale factor of 8}
\label{section:upscale8}
\begin{figure*}[!t]
    \centering
    \includegraphics[width=1\textwidth]{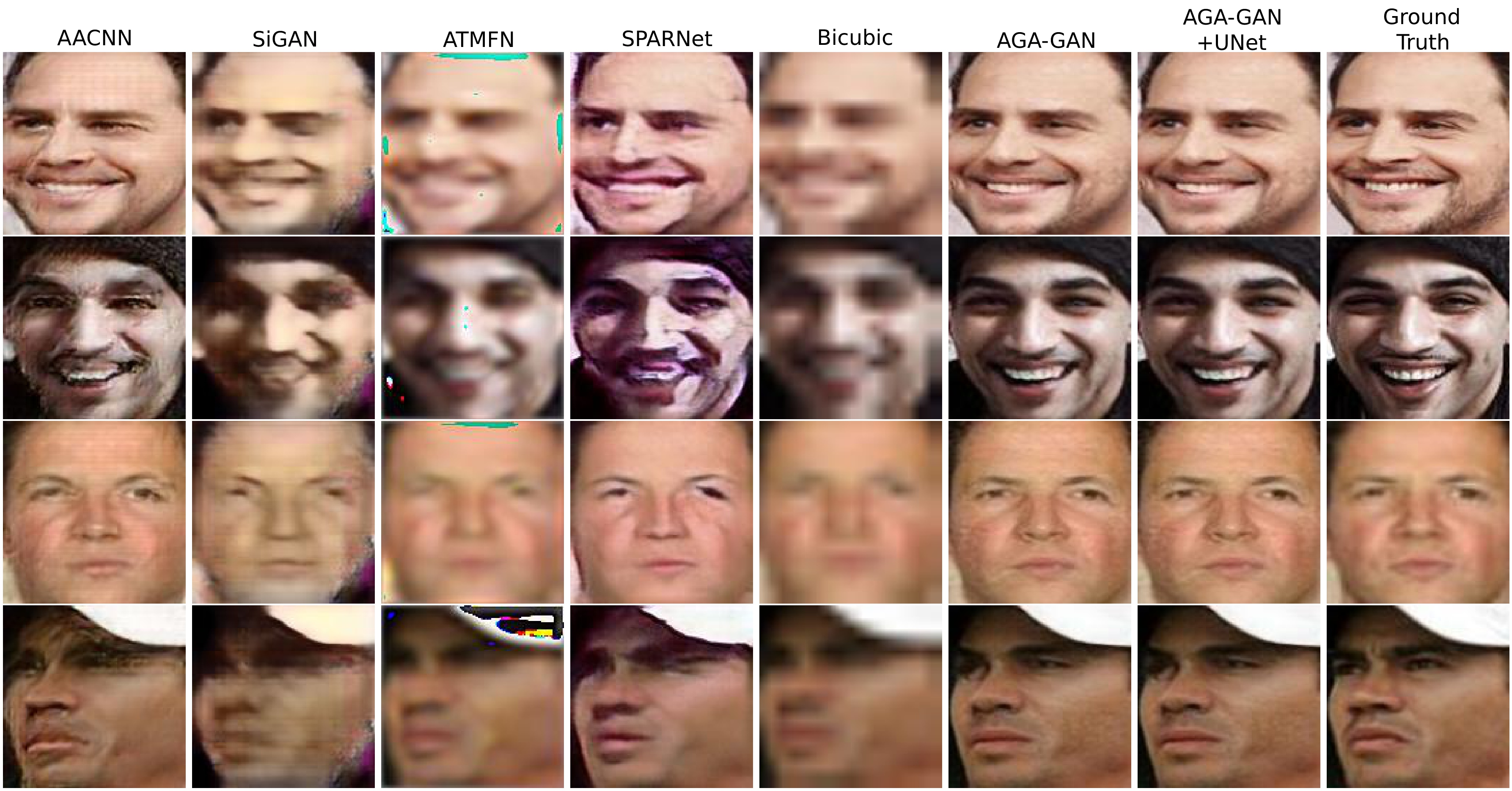}
    \caption{Qualitative Comparision of AGA-GAN with other SOTA methods on 8x upscaling}
    \label{fig:qualitative8}
\end{figure*}

In this experiment we aim to upscale the image by a factor of 8, the original 128 x 128 images are downsampled by a factor of 8 before feeding it into our AGA-GAN+U-Net framework. From quantitative results in Table~\ref{tab:result1} we can observe that AGA-GAN and AGA-GAN+U-Net outperform all other SOTA methods across all calculated metrics. The qualitative comparison with other methods on the CelebA dataset is shown in Figure~\ref{fig:qualitative8}. We can observe that bicubic interpolation of the image causes a major loss in facial details of the face whereas, SparNet with the help of spatial attention mechanism generates finer facial features. ATMFN effectively captures the facial structure but lacks detailed features causing the image to be blurry. AACNN can generate detailed and fine images but some artifacts persist. Overall our AGA-GAN and AGA-GAN+U-Net produce sharper images of superior visual quality, devoid of disturbing artifacts. Additionally, from Table~\ref{tab:result1} it can be observed that AGA-GAN and AGA-GAN+UNet has only 9.77M and 13.63M parameters which is significantly lower than SparNet~\cite{chen2020learning}, ATMFN~\cite{jiang2019atmfn} and comparable to SiGAN~\cite{hsu2019sigan}.

\begin{table}[!t]
%\centering
%\scriptsize
\begin{adjustwidth}{-1cm}{}
\footnotesize

\caption{Result comparison on the Celeb-A with an upscale ratio of 8}
\label{tab:result1}
\begin{tabular}{@{}l|l|l|l|l|l|l|l@{}}
\textbf{Method} & \textbf{PSNR} & \textbf{SSIM} & \textbf{FSIM} & \textbf{SRE} & \textbf{UIQ} & \textbf{BRISQUE} & \textbf{Parameters}\\ 
\hline
\hline
AGA-GAN (Ours) &\textcolor{blue}{31.4405} &\textcolor{blue}{0.7848} &\textcolor{blue}{0.6509}  &\textcolor{blue}{52.4967} &\textcolor{blue}{0.6679} &\textcolor{blue}{37.0216} &9.77M  \\ \hline
AGA-GAN+U-Net(Ours)&\textbf{31.9171} &\textbf{0.8149} &\textbf{0.6670}  &\textbf{52.8662} &\textbf{0.7079} &\textbf{31.2508} & 13.63M        \\ \hline
SparNet~\cite{chen2020learning} &29.2508 &0.6692 &0.5841  &50.2564 &0.5206 &43.1848 &25.44M        \\ \hline
ATMFN~\cite{jiang2019atmfn} &29.6547 &0.6367 &0.5492  &48.8089 &0.4727 &80.7350 & 64.26M      \\ \hline
SiGAN~\cite{hsu2019sigan} &28.9507 &0.6364 &0.5495  &49.0489 &0.4653 &42.2978 &13.38M       \\ \hline
AACNN~\cite{lee2018attribute} &30.2314 &0.6819 &0.5863  &51.2364 &0.5378& 38.2990 & 2.49M        \\ \hline
Bicubic &29.2394 &0.5568 &0.5137 &49.2655 &0.3573 & 89.6799 & -      \\ \hline

\end{tabular}

\end{adjustwidth}
\end{table}

\subsubsection{Comparison with the scale factor of 4}
\label{section:upscale4}
\begin{figure*}[!t]
    \centering
    \includegraphics[width=1\textwidth]{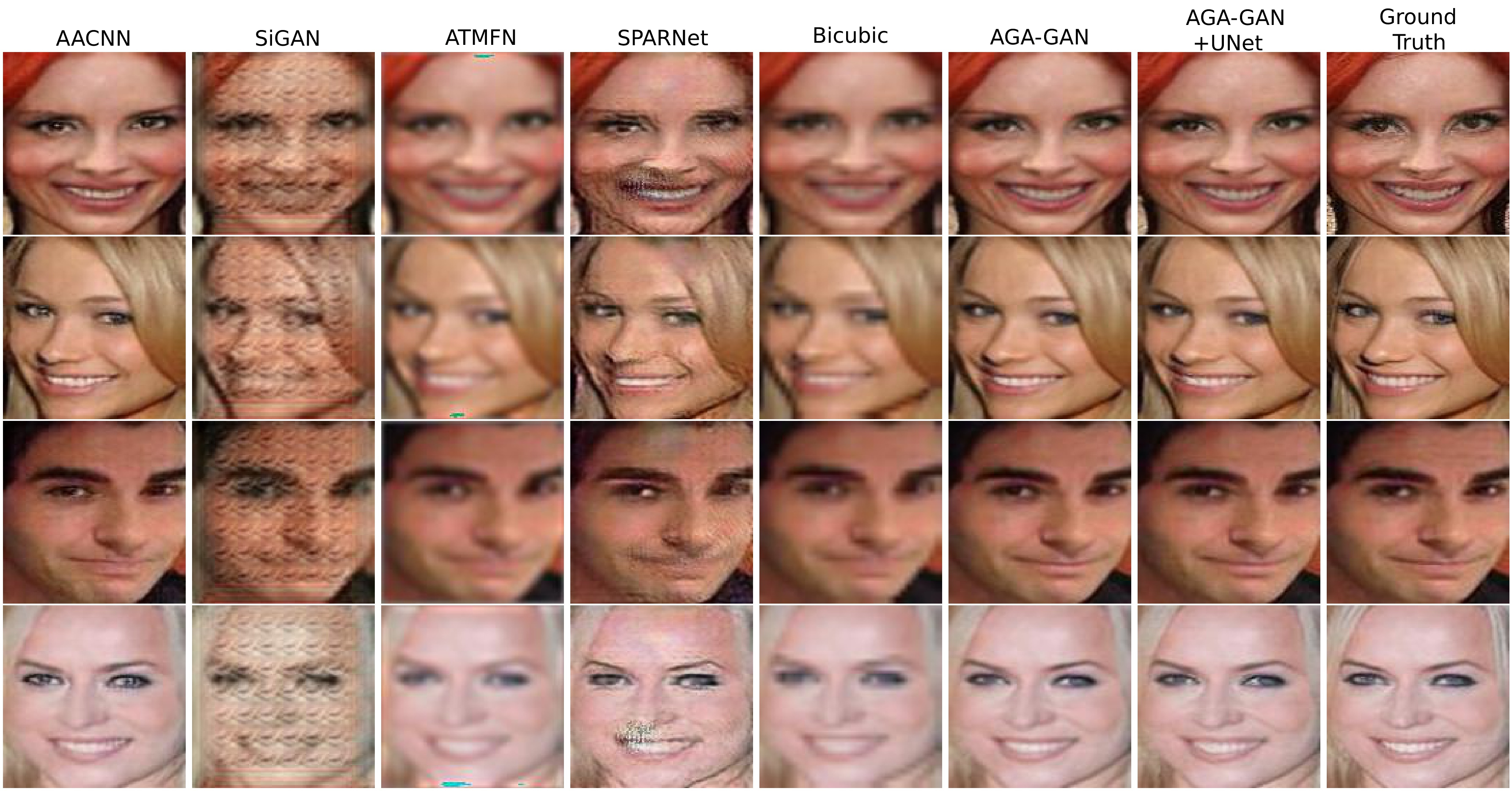}
    \caption{Qualitative Comparision of AGA-GAN with other SOTA methods on 4x upscaling}
    \label{fig:qualitative4}
\end{figure*}

In this experiment, we aim to upscale the image by a factor of 4, the original 128 x 128 images are downsampled by a factor of 4 to 32 x 32 before it is coupled with its corresponding attribute vector and served as input to our model. The quantitative results are provided in Table~\ref{tab:result4x} and the qualitative results are shown in Figure~\ref{fig:qualitative4}. We can observe from Table~\ref{tab:result4x} that AGA-GAN+U-Net surpasses all other SOTA methods in all metrics. In Figure~\ref{fig:qualitative4} we can observe that each method is effective to recover face structures to an extent. AACNN can generate fine details but noise and irregularities in the facial structures are observed. SiGAN and ATMFN synthesize blurry and fuzzy predictions. Although SPARNet demonstrates decent capacity in recovering sharper features which can be attributed to its facial attention units, the predictions are susceptible to noises. Bicubic interpolation lacks the desired feature details in its predictions. AGA-GAN and AGA-GAN+U-Net exhibit superior performance by generating high-quality SR predictions. The images accurately capture the facial structures and are enriched with fine and intricate details and consequently possess a high degree of fidelity. 

\begin{table}[!t]
%\centering
%\scriptsize
\begin{adjustwidth}{-1cm}{}
\footnotesize
\caption{Result comparison on the Celeb-A with an upscale ratio of 4}
\label{tab:result4x}
\begin{tabular}{@{}l|l|l|l|l|l|l|l@{}}
\textbf{Method} & \textbf{PSNR} & \textbf{SSIM} & \textbf{FSIM} & \textbf{SRE} & \textbf{UIQ} &\textbf{BRISQUE} & \textbf{Parameters}\\ 
\hline
\hline
AGA-GAN (Ours)&\textcolor{blue}{34.1323} &\textcolor{blue}{0.9104} &
\textcolor{blue}{0.7522}  &\textcolor{blue}{54.9578} &\textcolor{blue}{0.8630} &\textbf{33.7918} & 9.26M  \\ \hline
AGA-GAN+U-Net(Ours) &\textbf{34.7659} &\textbf{0.9247} &\textbf{0.7676}  &\textbf{55.3552} &\textbf{0.8848} &\textcolor{blue}{36.0212} &13.12M        \\ \hline
SparNet~\cite{chen2020learning} &30.3309 &0.7584 &0.6382  &51.7745 &0.6497 &40.6698 &25.44M       \\ \hline
ATMFN~\cite{jiang2019atmfn} &30.7626 &0.7624 &0.6197  &49.7543 &0.6715 &70.2202 & 78.21M        \\ \hline
SiGAN~\cite{hsu2019sigan} &29.1226 &0.5881 &0.5749  &50.5577 &0.4783 &39.5480 &3.27M      \\ \hline
AACNN~\cite{lee2018attribute} &32.4497 &0.8703 &0.7104  &53.6478 &0.7892 &43.8673 &2.34M        \\ \hline
Bicubic &30.6845 &0.7513 &0.6315 &51.2223 &0.6463 & 60.2453 & -      \\ \hline

\end{tabular}

\end{adjustwidth}
\end{table}
%%%%%%%%%%%%%%%%%%%%%%%% NEW SECTION

%%%%%%%%%%%%%%%%%%%%%%%%%%
\subsection{Evaluation on Attribute Specific Facial Images}
\label{section:local_region}
\begin{figure*}[!t]
    \centering
    \includegraphics[width=1\textwidth]{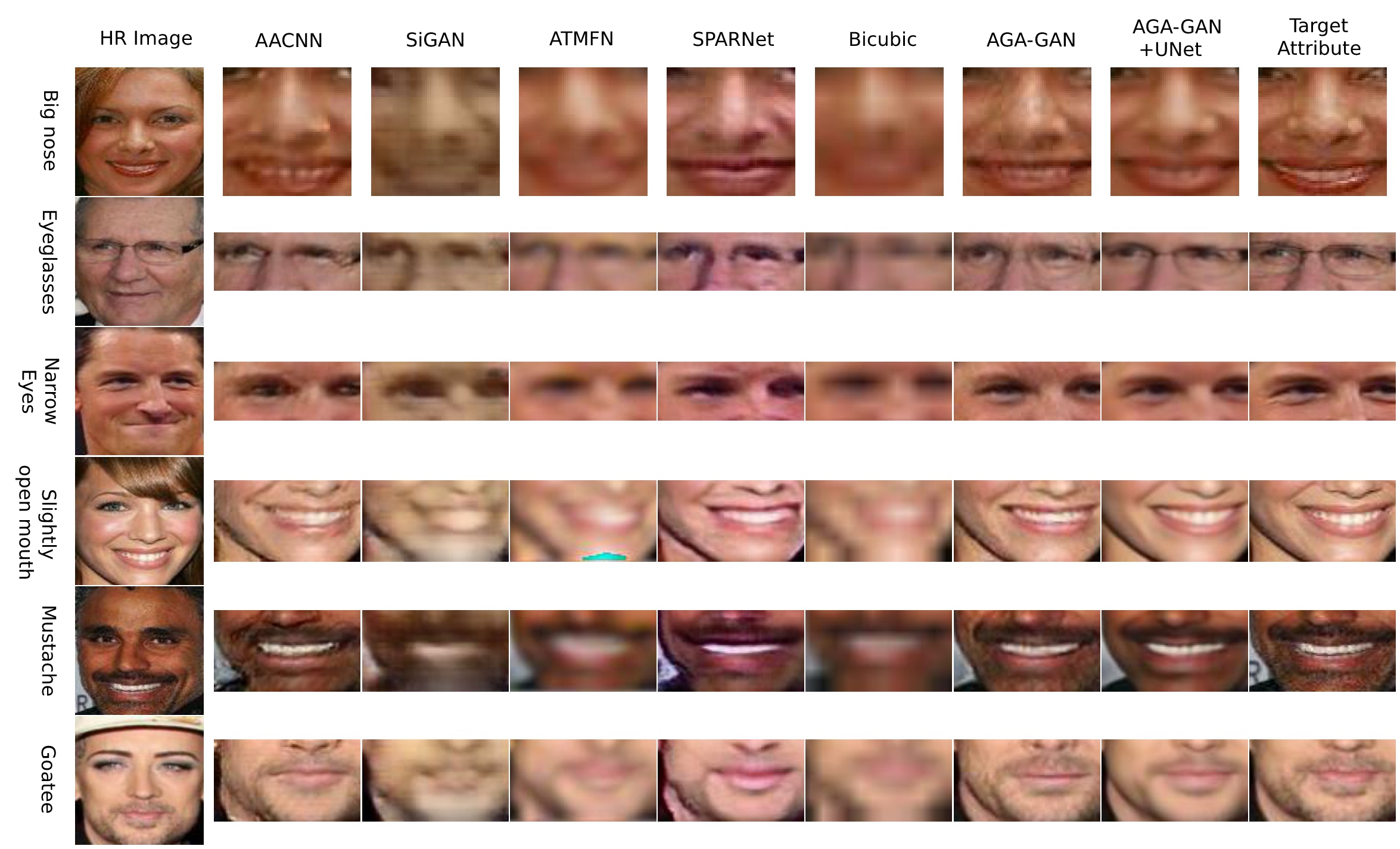}
    \caption{Qualitative Comparision of AGA-GAN+U-Net with other SOTA methods on specific facial attributes}
    \label{fig:local}
\end{figure*}

In this section, we study the performance of our selected methods against our AGA-GAN+U-Net framework on subsets of the Celeb-A dataset which contains specific attributes. We present both quantitative and qualitative comparisons of each method's competency in recovering facial attributes from the input LR image. We ensure that these subsets maintain maximum overlap with the original test set on the 8x upscaling task. The detailed qualitative comparison is presented in Figure~\ref{fig:local}. We choose the attributes \enquote{big nose}, \enquote{eyeglasses}, \enquote{goatee}, \enquote{slightly open mouth}, \enquote{mustache} and \enquote{narrow eyes} as our primary facial attributes to be recovered in this experiment. These attributes cover the eye, nose, and mouth region of the face and possess a varying amount of difficulty in recovery.
In Table~\ref{tab:resultbignose} we report the quantitative comparison of AGA-GAN+U-Net framework with other methods on images that have \enquote{big nose}. We can observe that the performance of all methods drops as compared to the global 8x upscaling experiment(see Table~\ref{tab:result1}) except for ATMFN whose performance improves. AGA-GAN+U-Net reports the best performance across all metrics in this category. In Figure~\ref{fig:local} we can observe that each method can recover and determine the size of the nose to an extent.  We notice that AGA-GAN and AGA-GAN+U-Net are effective in recovering structures and detailed features of the target attribute. 

\begin{table}[!t]
\centering
\begin{adjustwidth}{-1cm}{}
%\scriptsize
\footnotesize
\caption{Result comparison of big nose attribute}
\label{tab:resultbignose}
\begin{tabular}{@{}l|l|l|l|l|l|l|l@{}}
\toprule
\textbf{Method} & \textbf{PSNR} & \textbf{SSIM} & \textbf{FSIM} & \textbf{SRE} & \textbf{UIQ}& \textbf{BRISQUE} & \textbf{Parameters}\\ 
\hline
\hline
AGA-GAN(Ours) &\textcolor{blue}{31.0675} &\textcolor{blue}{0.7519} &\textcolor{blue}{0.6309}  &\textcolor{blue}{51.6057} &\textcolor{blue}{0.6378} &\textcolor{blue}{36.1596} & 9.77M \\ \hline
AGA-GAN+U-Net(Ours) &\textbf{31.5187} &\textbf{0.7848} &\textbf{0.6446}  &\textbf{51.9691} &\textbf{0.6777} &\textbf{30.3936} &13.63M       \\ \hline
SparNet~\cite{chen2020learning} &29.2007 &0.6407 &0.5690  &49.5539 &0.5010 &43.7745 &25.44M       \\ \hline
ATMFN~\cite{jiang2019atmfn} &29.7448 &0.6163 &0.5411  &48.2906 &0.4634 &80.7797 & 64.26M        \\ \hline
SiGAN~\cite{hsu2019sigan} &28.8972 &0.6067 &0.5353  &48.3891 &0.4422 &42.2193 &13.38M      \\ \hline
AACNN~\cite{lee2018attribute} &30.0141 &0.6450 &0.5676  &50.4107 &0.5078 &36.6382 &2.49M        \\ \hline
Bicubic &29.1823 &0.5314 &0.5027 &48.6661 &0.3353 & 89.5779 & -    \\ \hline
\bottomrule
\end{tabular}

\end{adjustwidth}
\end{table}

Table~\ref{tab:resulteyeglasses} and Table~\ref{tab:resultnarrow_eyes} report the performance of each method on images that possess eyeglasses and have narrow eyes as facial features, respectively. We can observe that these attributes are relatively hard to recover with faces having eyeglasses reporting the lowest results across all metrics and methods as compared to other attributes. This is also supported by Figure~\ref{fig:local} where we can see that most methods are unable to generate the eyeglasses. This can be attributed to the fact that the LR image is 16 x 16, which may contain a very limited amount of information regarding eyeglasses. Even though attribute descriptors relay the information about the presence of eyeglasses to AACNN's network, it still fails to recover them. AGA-GAN and AGA-GAN+U-Net, leveraging the attribute guided attention module is effective in recovering this attribute. While AGA-GAN lacks the finer details, AGA-GAN+U-Net can further refine and generate sharper and visually pleasing details of the eyeglasses. 
Another region we explore is the mouth area of the facial images. Attributes that describe the facial hair of the image i.e. \enquote{goatee} and \enquote{mustache} are hardest to recover in this region as evident in Table~\ref{tab:resultgoatee} and Table~\ref{tab:resultmustache} whereas, \enquote{slightly open mouth} is effectively recovered as shown in Table~\ref{tab:resultmouth_open}. In Figure~\ref{fig:local} we can note that even though these attributes are prominent in each method's predictions, they lack the finer details present in the original HR image.

\begin{table}[!t]
\centering
\begin{adjustwidth}{-1cm}{}
%\scriptsize
\footnotesize
\caption{Result comparison narrow eyes attribute}
\label{tab:resultnarrow_eyes}
\begin{tabular}{@{}l|l|l|l|l|l|l|l@{}}
\toprule
\textbf{Method} & \textbf{PSNR} & \textbf{SSIM} & \textbf{FSIM} & \textbf{SRE} & \textbf{UIQ}& \textbf{BRISQUE} & \textbf{Parameters}\\ 
\hline
\hline
AGA-GAN (Ours) &\textcolor{blue}{31.3206} &\textcolor{blue}{0.7790} &\textcolor{blue}{0.6482}  &\textcolor{blue}{52.4059} &\textcolor{blue}{0.6634} &38.8545 & 9.77M  \\ \hline
AGA-GAN+U-Net(Ours) &\textbf{31.7804} &\textbf{0.8090} &\textbf{0.6642}  &\textbf{52.7705} &\textbf{0.7028} &\textbf{32.2972} &13.63M        \\ \hline
SparNet~\cite{chen2020learning} &29.2244 &0.6623 &0.5821  &50.1950 &0.5156 &34.6491 &25.44M       \\ \hline
ATMFN~\cite{jiang2019atmfn} &29.6100 &0.6301 &0.5478  &48.7321 &0.4676 &80.1130 & 64.26M        \\ \hline
SiGAN~\cite{hsu2019sigan} &28.9139 &0.6311 &0.5500  &49.0028 &0.4619 &42.7260 &13.38M      \\ \hline
AACNN~\cite{lee2018attribute} &30.1428 &0.6732 &0.5828  &51.1525 &0.5306 &\textcolor{blue}{38.7267} & 2.49M        \\ \hline
Bicubic &29.2066 &0.5480 &0.5131 &49.2006 &0.3500 &89.3715 & -     \\ \hline
\bottomrule

\end{tabular}

\end{adjustwidth}
\end{table}

\begin{table}[!t]
\centering
\begin{adjustwidth}{-1cm}{}
%\scriptsize
\footnotesize

\caption{Result comparison of eyeglasses attribute}
\label{tab:resulteyeglasses}
\begin{tabular}{@{}l|l|l|l|l|l|l|l@{}}
\toprule
\textbf{Method} & \textbf{PSNR} & \textbf{SSIM} & \textbf{FSIM} & \textbf{SRE} & \textbf{UIQ}& \textbf{BRISQUE} & \textbf{Parameters}\\ 
\hline
\hline
AGA-GAN (Ours) &\textcolor{blue}{30.7578} &\textcolor{blue}{0.7122} &\textcolor{blue}{0.6104}  &\textcolor{blue}{50.9109} &\textcolor{blue}{0.5882} &\textcolor{blue}{36.9800} & 9.77M  \\ \hline
AGA-GAN+U-Net(Ours) &\textbf{31.1469} &\textbf{0.7468} &\textbf{0.6245}  &\textbf{51.2552} &\textbf{0.6298} &\textbf{30.9812}  &13.63M       \\ \hline
SparNet~\cite{chen2020learning} &29.1150 &0.5967 &0.5520  &49.0468 &0.4537 &34.9088  &25.44M       \\ \hline
ATMFN~\cite{jiang2019atmfn} &29.4592 &0.5755 &0.5203  &47.7411 &0.4184 &78.0582 &64.26M        \\ \hline
SiGAN~\cite{hsu2019sigan} &28.8148 &0.5818 &0.5288  &48.1265 &0.4108 &43.8321 &13.38M      \\ \hline
AACNN~\cite{lee2018attribute} &29.8511 &0.6096 &0.5547  &49.9227 &0.4662  &39.9565 &2.49M      \\ \hline
Bicubic &29.1504 &0.5072 &0.4947 &48.2397 &0.3146 & 89.2502 & -     \\ \hline
\bottomrule

\end{tabular}

\end{adjustwidth}
\end{table}

\begin{table}[!t]
\centering
\begin{adjustwidth}{-1cm}{}
%\scriptsize
\footnotesize

\caption{Result comparison of goatee attribute}
\label{tab:resultgoatee}
\begin{tabular}{@{}l|l|l|l|l|l|l|l@{}}
\toprule
\textbf{Method} & \textbf{PSNR} & \textbf{SSIM} & \textbf{FSIM} & \textbf{SRE} & \textbf{UIQ}& \textbf{BRISQUE} & \textbf{Parameters}\\ 
\hline
\hline
AGA-GAN (Ours) &\textcolor{blue}{30.8354} &\textcolor{blue}{0.7333} &\textcolor{blue}{0.61758}  &\textcolor{blue}{50.9577} &\textcolor{blue}{0.6232} &37.0496 & 9.77M  \\ \hline
AGA-GAN+U-Net(Ours) &\textbf{31.2479} &\textbf{0.7663} &\textbf{0.6309}  &\textbf{51.3081} &\textbf{0.6625} &\textbf{31.0462}  &13.63M     \\ \hline
SparNet~\cite{chen2020learning} &29.1498 &0.6207 &0.5570  &48.9490 &0.4875 &33.8498 &25.44M       \\ \hline
ATMFN~\cite{jiang2019atmfn} &29.4613 &0.5976 &0.5311  &47.7364 &0.4490 &81.2881 &64.26M      \\ \hline
SiGAN~\cite{hsu2019sigan} &28.8812 &0.5942 &0.5302  &47.9543 &0.4354 &42.6179 &13.38M    \\ \hline
AACNN~\cite{lee2018attribute} &29.8320 &0.6151 &0.5512  &49.7123 &0.4840 &\textcolor{blue}{32.3641} &2.49M       \\ \hline
Bicubic &29.1275 &0.5209 &0.4989 &48.1002 &0.3360 &89.5204 & -     \\ \hline
\bottomrule

\end{tabular}

\end{adjustwidth}
\end{table}

\begin{table}[!t]
\centering
\begin{adjustwidth}{-1cm}{}
%\scriptsize
\footnotesize

\caption{Result comparison of mouth open attribute}
\label{tab:resultmouth_open}
\begin{tabular}{@{}l|l|l|l|l|l|l|l@{}}
\toprule
\textbf{Method} & \textbf{PSNR} & \textbf{SSIM} & \textbf{FSIM} & \textbf{SRE} & \textbf{UIQ}& \textbf{BRISQUE} & \textbf{Parameters}\\ 
\hline
\hline
AGA-GAN (Ours) &\textcolor{blue}{31.3236} &\textcolor{blue}{0.7799} &\textcolor{blue}{0.6492}  &\textcolor{blue}{52.4407} &\textcolor{blue}{0.6685} &\textcolor{blue}{36.3803} & 9.77M \\ \hline
AGA-GAN+U-Net(Ours) &\textbf{31.7866} &\textbf{0.8102} &\textbf{0.6653}  &\textbf{52.8103} &\textbf{0.7081} &\textbf{30.1729} &13.63M        \\ \hline
SparNet~\cite{chen2020learning} &29.2162 &0.6650 &0.5834  &50.2908 &0.5224 &33.9014 &25.44M     \\ \hline
ATMFN~\cite{jiang2019atmfn} &29.5945 &0.6299 &0.5493  &48.9114 &0.4691 &80.1840 &64.26M        \\ \hline
SiGAN~\cite{hsu2019sigan} &28.9329 &0.6276 &0.5471  &49.0956 &0.4620 &42.1772 &13.38M      \\ \hline
AACNN~\cite{lee2018attribute} &30.1581 &0.6744 &0.5839  &51.2004 &0.5365 &37.8457 &2.49M     \\ \hline
Bicubic &29.1911 &0.5462 &0.5108 &49.2990 &0.3504& 89.4186 & -     \\ \hline
\bottomrule

\end{tabular}

\end{adjustwidth}
\end{table}

\begin{table}[!t]
\centering
\begin{adjustwidth}{-1cm}{}
%\scriptsize
\footnotesize

\caption{Result comparison mustache attribute}
\label{tab:resultmustache}
\begin{tabular}{@{}l|l|l|l|l|l|l|l@{}}
\toprule
\textbf{Method} & \textbf{PSNR} & \textbf{SSIM} & \textbf{FSIM} & \textbf{SRE} & \textbf{UIQ}& \textbf{BRISQUE} & \textbf{Parameters}\\ 
\hline
\hline
AGA-GAN (Ours)&\textcolor{blue}{30.8353} &\textcolor{blue}{0.7323} &\textcolor{blue}{0.6180}  &\textcolor{blue}{50.8118} &\textcolor{blue}{0.6216} &37.7595 & 9.77M  \\ \hline
AGA-GAN+U-Net (Ours)&\textbf{31.2548} &\textbf{0.7655} &\textbf{0.6313}  &\textbf{51.1707} &\textbf{0.6608} &\textbf{31.5656} &13.63M        \\ \hline
SparNet~\cite{chen2020learning} &29.1402 &0.6201 &0.5581  &48.7875 &0.4874 &34.0678 &25.44M      \\ \hline
ATMFN~\cite{jiang2019atmfn} &29.4444 &0.5964 &0.5321  &47.5626 &0.4467 &81.1625 &64.26M      \\ \hline
SiGAN~\cite{hsu2019sigan} &28.8599 &0.5915 &0.5298  &47.7583 &0.4317 &42.5275 &13.38M    \\ \hline
AACNN~\cite{lee2018attribute} &29.8099 &0.6105 &0.5504  &49.5380 &0.4798 &\textcolor{blue}{31.5483} &2.49M       \\ \hline
Bicubic &29.1158 &0.5186 &0.4992 &47.9362 &0.3318 & 89.5672 & -     \\ \hline
\bottomrule

\end{tabular}

\end{adjustwidth}
\end{table}

\subsection{Effect of Feature Refinement using U-Net}
\label{section:refinementU-Net}
\begin{figure*}[!t]
\centering
    \includegraphics[width=0.7\textwidth,height=0.7\textwidth]{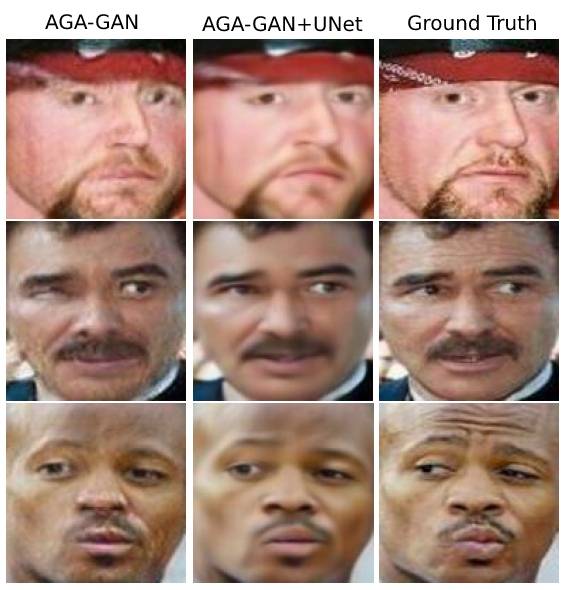}
    \caption{Feature refinement using U-Net}
    \label{fig:U-Net_q}
\end{figure*}

We study the impact of adding a spatial and channel attention U-Net on the existing AGA-GAN. Using the prediction of AGA-GAN as input for U-Net, the network works on improving the facial features and generating finer and richer details to improve the overall visual quality. Performance gain observed from Table~\ref{tab:result1} and Table~\ref{tab:result4x} show that U-Net is successful in refining and improving the quality of existing SR predictions by AGA-GAN. In Figure~\ref{fig:U-Net_q} we can observe the various artifacts generated in the SR predictions of AGA-GAN and their subsequent rectification and refinement by U-Net. This proves the scope of using U-Net for image refinement of candidate predictions by other face hallucination methods.

\subsection{Evaluation when Attribute information is scarce}
\label{section:scarceatt}
In real-world scenarios, it is not feasible that entire attribute information will be attainable with each LR image. To demonstrate the performance of AGA-GAN in the case when only partial attribute information is present we represent some known attributes is unknown. In Table~\ref{tab:resultunkown} we observe that when 50\% of the attributes are known, a drop in performance is noted which further drops when 75\% of the attributes are unknown. The performance is still superior to other methods and demonstrates that AGA-GAN can be used in situations when only partial attribute information is present.

\begin{table}[H]
\centering
%\scriptsize
\footnotesize
\caption{Result comparison with partial unknown attribute sets}
\label{tab:resultunkown}
\begin{tabular}{@{}l|l|l|l|l|l@{}}
\toprule
\textbf{Unknown attributes} & \textbf{PSNR} & \textbf{SSIM} & \textbf{FSIM} & \textbf{SRE} & \textbf{UIQ}\\ 
\hline
\hline
50\% &31.4353 &0.7839 &0.6496  &52.4684 &0.6670  \\ \hline
75\% &31.4266 &0.7801 &0.6450  &52.4356 &0.6661        \\ \hline
\bottomrule

\end{tabular}

\end{table}
\section{Conclusion}
\label{section:conclusion}
In this work, we have presented our attribute-guided attention generative adversarial network (AGA-GAN) which utilizes attributes before an LR image to generate attention maps capable of progressively generating richer and spatially accurate high-resolution feature maps, eventually generating high-resolution predictions with high fidelity and rich features. We also explore the use of spatial and channel attention U-Net for refining and generating additional facial features in an existing SR prediction. Extensive experiments with five metrics display the superior performance of both AGA-GAN and AGA-GAN+U-Net for facial hallucination. Additionally, experiments with partially known attributes demonstrate the practicality of our method in real-world scenarios. Our future work will comprise of extending our AGA-GAN+U-Net framework for face anti-spoofing and face parsing. 
\section{Acknowledgement}
This is a collaborative research work between Indian Statistical Institute, Kolkata, India and Østfold University College, Halden, Norway. The experiments in this paper were performed on a high performance computing platform \enquote{Experimental Infrastructure for Exploration of Exascale Computing} (eX3), which is funded by the Research Council of Norway.

\bibliographystyle{elsarticle-num-names} 
\bibliography{refer}

\begin{thebibliography}{44}
\expandafter\ifx\csname natexlab\endcsname\relax\def\natexlab#1{#1}\fi
\providecommand{\url}[1]{\texttt{#1}}
\providecommand{\href}[2]{#2}
\providecommand{\path}[1]{#1}
\providecommand{\DOIprefix}{doi:}
\providecommand{\ArXivprefix}{arXiv:}
\providecommand{\URLprefix}{URL: }
\providecommand{\Pubmedprefix}{pmid:}
\providecommand{\doi}[1]{\href{http://dx.doi.org/#1}{\path{#1}}}
\providecommand{\Pubmed}[1]{\href{pmid:#1}{\path{#1}}}
\providecommand{\bibinfo}[2]{#2}
\ifx\xfnm\relax \def\xfnm[#1]{\unskip,\space#1}\fi
%Type = Article
\bibitem[{Zou and Yuen(2011)}]{zou2011very}
\bibinfo{author}{W.~W. Zou}, \bibinfo{author}{P.~C. Yuen},
\newblock \bibinfo{title}{Very low resolution face recognition problem},
\newblock \bibinfo{journal}{IEEE Transactions on image processing}
  \bibinfo{volume}{21} (\bibinfo{year}{2011}) \bibinfo{pages}{327--340}.
%Type = Article
\bibitem[{Ma et~al.(2010)Ma, Zhang, and Qi}]{ma2010hallucinating}
\bibinfo{author}{X.~Ma}, \bibinfo{author}{J.~Zhang}, \bibinfo{author}{C.~Qi},
\newblock \bibinfo{title}{Hallucinating face by position-patch},
\newblock \bibinfo{journal}{Pattern Recognition} \bibinfo{volume}{43}
  (\bibinfo{year}{2010}) \bibinfo{pages}{2224--2236}.
%Type = Article
\bibitem[{Zhang and Wu(2008)}]{zhang2008image}
\bibinfo{author}{X.~Zhang}, \bibinfo{author}{X.~Wu},
\newblock \bibinfo{title}{Image interpolation by adaptive 2-d autoregressive
  modeling and soft-decision estimation},
\newblock \bibinfo{journal}{IEEE transactions on image processing}
  \bibinfo{volume}{17} (\bibinfo{year}{2008}) \bibinfo{pages}{887--896}.
%Type = Article
\bibitem[{Sun et~al.(2010)Sun, Xu, and Shum}]{sun2010gradient}
\bibinfo{author}{J.~Sun}, \bibinfo{author}{Z.~Xu}, \bibinfo{author}{H.-Y.
  Shum},
\newblock \bibinfo{title}{Gradient profile prior and its applications in image
  super-resolution and enhancement},
\newblock \bibinfo{journal}{IEEE Transactions on Image Processing}
  \bibinfo{volume}{20} (\bibinfo{year}{2010}) \bibinfo{pages}{1529--1542}.
%Type = Article
\bibitem[{Zhang et~al.(2015)Zhang, Tao, Gao, Li, and Xiong}]{zhang2015learning}
\bibinfo{author}{K.~Zhang}, \bibinfo{author}{D.~Tao}, \bibinfo{author}{X.~Gao},
  \bibinfo{author}{X.~Li}, \bibinfo{author}{Z.~Xiong},
\newblock \bibinfo{title}{Learning multiple linear mappings for efficient
  single image super-resolution},
\newblock \bibinfo{journal}{IEEE Transactions on Image Processing}
  \bibinfo{volume}{24} (\bibinfo{year}{2015}) \bibinfo{pages}{846--861}.
%Type = Inproceedings
\bibitem[{Chang et~al.(2004)Chang, Yeung, and Xiong}]{chang2004super}
\bibinfo{author}{H.~Chang}, \bibinfo{author}{D.-Y. Yeung},
  \bibinfo{author}{Y.~Xiong},
\newblock \bibinfo{title}{Super-resolution through neighbor embedding},
\newblock in: \bibinfo{booktitle}{Proceedings of the 2004 IEEE Computer Society
  Conference on Computer Vision and Pattern Recognition, 2004. CVPR 2004.},
  volume~\bibinfo{volume}{1}, \bibinfo{organization}{IEEE},
  \bibinfo{year}{2004}, pp. \bibinfo{pages}{I--I}.
%Type = Inproceedings
\bibitem[{Dong et~al.(2016)Dong, Loy, and Tang}]{dong2016accelerating}
\bibinfo{author}{C.~Dong}, \bibinfo{author}{C.~C. Loy},
  \bibinfo{author}{X.~Tang},
\newblock \bibinfo{title}{Accelerating the super-resolution convolutional
  neural network},
\newblock in: \bibinfo{booktitle}{European conference on computer vision},
  \bibinfo{organization}{Springer}, \bibinfo{year}{2016}, pp.
  \bibinfo{pages}{391--407}.
%Type = Inproceedings
\bibitem[{Shi et~al.(2016)Shi, Caballero, Husz{\'a}r, Totz, Aitken, Bishop,
  Rueckert, and Wang}]{shi2016real}
\bibinfo{author}{W.~Shi}, \bibinfo{author}{J.~Caballero},
  \bibinfo{author}{F.~Husz{\'a}r}, \bibinfo{author}{J.~Totz},
  \bibinfo{author}{A.~P. Aitken}, \bibinfo{author}{R.~Bishop},
  \bibinfo{author}{D.~Rueckert}, \bibinfo{author}{Z.~Wang},
\newblock \bibinfo{title}{Real-time single image and video super-resolution
  using an efficient sub-pixel convolutional neural network},
\newblock in: \bibinfo{booktitle}{Proceedings of the IEEE conference on
  computer vision and pattern recognition}, \bibinfo{year}{2016}, pp.
  \bibinfo{pages}{1874--1883}.
%Type = Inproceedings
\bibitem[{Yu and Porikli(2016)}]{yu2016ultra}
\bibinfo{author}{X.~Yu}, \bibinfo{author}{F.~Porikli},
\newblock \bibinfo{title}{Ultra-resolving face images by discriminative
  generative networks},
\newblock in: \bibinfo{booktitle}{European conference on computer vision},
  \bibinfo{organization}{Springer}, \bibinfo{year}{2016}, pp.
  \bibinfo{pages}{318--333}.
%Type = Article
\bibitem[{Hsu et~al.(2019)Hsu, Lin, Su, and Cheung}]{hsu2019sigan}
\bibinfo{author}{C.-C. Hsu}, \bibinfo{author}{C.-W. Lin},
  \bibinfo{author}{W.-T. Su}, \bibinfo{author}{G.~Cheung},
\newblock \bibinfo{title}{Sigan: Siamese generative adversarial network for
  identity-preserving face hallucination},
\newblock \bibinfo{journal}{IEEE Transactions on Image Processing}
  \bibinfo{volume}{28} (\bibinfo{year}{2019}) \bibinfo{pages}{6225--6236}.
%Type = Inproceedings
\bibitem[{Ronneberger et~al.(2015)Ronneberger, Fischer, and
  Brox}]{ronneberger2015u}
\bibinfo{author}{O.~Ronneberger}, \bibinfo{author}{P.~Fischer},
  \bibinfo{author}{T.~Brox},
\newblock \bibinfo{title}{U-net: Convolutional networks for biomedical image
  segmentation},
\newblock in: \bibinfo{booktitle}{International Conference on Medical image
  computing and computer-assisted intervention},
  \bibinfo{organization}{Springer}, \bibinfo{year}{2015}, pp.
  \bibinfo{pages}{234--241}.
%Type = Inproceedings
\bibitem[{Baker and Kanade(2000)}]{baker2000hallucinating}
\bibinfo{author}{S.~Baker}, \bibinfo{author}{T.~Kanade},
\newblock \bibinfo{title}{Hallucinating faces},
\newblock in: \bibinfo{booktitle}{Proceedings Fourth IEEE international
  conference on automatic face and gesture recognition (Cat. No. PR00580)},
  \bibinfo{organization}{IEEE}, \bibinfo{year}{2000}, pp.
  \bibinfo{pages}{83--88}.
%Type = Inproceedings
\bibitem[{Liu et~al.(2001)Liu, Shum, and Zhang}]{liu2001two}
\bibinfo{author}{C.~Liu}, \bibinfo{author}{H.-Y. Shum}, \bibinfo{author}{C.-S.
  Zhang},
\newblock \bibinfo{title}{A two-step approach to hallucinating faces: global
  parametric model and local nonparametric model},
\newblock in: \bibinfo{booktitle}{Proceedings of the 2001 IEEE Computer Society
  Conference on Computer Vision and Pattern Recognition. CVPR 2001},
  volume~\bibinfo{volume}{1}, \bibinfo{organization}{IEEE},
  \bibinfo{year}{2001}, pp. \bibinfo{pages}{I--I}.
%Type = Article
\bibitem[{Gunturk et~al.(2003)Gunturk, Batur, Altunbasak, Hayes, and
  Mersereau}]{gunturk2003eigenface}
\bibinfo{author}{B.~K. Gunturk}, \bibinfo{author}{A.~U. Batur},
  \bibinfo{author}{Y.~Altunbasak}, \bibinfo{author}{M.~H. Hayes},
  \bibinfo{author}{R.~M. Mersereau},
\newblock \bibinfo{title}{Eigenface-domain super-resolution for face
  recognition},
\newblock \bibinfo{journal}{IEEE transactions on image processing}
  \bibinfo{volume}{12} (\bibinfo{year}{2003}) \bibinfo{pages}{597--606}.
%Type = Article
\bibitem[{Wang and Tang(2005)}]{wang2005hallucinating}
\bibinfo{author}{X.~Wang}, \bibinfo{author}{X.~Tang},
\newblock \bibinfo{title}{Hallucinating face by eigentransformation},
\newblock \bibinfo{journal}{IEEE Transactions on Systems, Man, and Cybernetics,
  Part C (Applications and Reviews)} \bibinfo{volume}{35}
  (\bibinfo{year}{2005}) \bibinfo{pages}{425--434}.
%Type = Article
\bibitem[{Liang et~al.(2013)Liang, Xie, and Lai}]{liang2013face}
\bibinfo{author}{Y.~Liang}, \bibinfo{author}{X.~Xie}, \bibinfo{author}{J.-H.
  Lai},
\newblock \bibinfo{title}{Face hallucination based on morphological component
  analysis},
\newblock \bibinfo{journal}{Signal Processing} \bibinfo{volume}{93}
  (\bibinfo{year}{2013}) \bibinfo{pages}{445--458}.
%Type = Inproceedings
\bibitem[{Zhou et~al.(2015)Zhou, Fan, Cao, Jiang, and Yin}]{zhou2015learning}
\bibinfo{author}{E.~Zhou}, \bibinfo{author}{H.~Fan}, \bibinfo{author}{Z.~Cao},
  \bibinfo{author}{Y.~Jiang}, \bibinfo{author}{Q.~Yin},
\newblock \bibinfo{title}{Learning face hallucination in the wild},
\newblock in: \bibinfo{booktitle}{Proceedings of the AAAI Conference on
  Artificial Intelligence}, volume~\bibinfo{volume}{29}, \bibinfo{year}{2015}.
%Type = Article
\bibitem[{Chen et~al.(2020)Chen, Wang, Lu, Li, Wang, and
  Huang}]{chen2020rbpnet}
\bibinfo{author}{X.~Chen}, \bibinfo{author}{X.~Wang}, \bibinfo{author}{Y.~Lu},
  \bibinfo{author}{W.~Li}, \bibinfo{author}{Z.~Wang},
  \bibinfo{author}{Z.~Huang},
\newblock \bibinfo{title}{Rbpnet: An asymptotic residual back-projection
  network for super-resolution of very low-resolution face image},
\newblock \bibinfo{journal}{Neurocomputing} \bibinfo{volume}{376}
  (\bibinfo{year}{2020}) \bibinfo{pages}{119--127}.
%Type = Inproceedings
\bibitem[{Huang et~al.(2017)Huang, He, Sun, and Tan}]{huang2017wavelet}
\bibinfo{author}{H.~Huang}, \bibinfo{author}{R.~He}, \bibinfo{author}{Z.~Sun},
  \bibinfo{author}{T.~Tan},
\newblock \bibinfo{title}{Wavelet-srnet: A wavelet-based cnn for multi-scale
  face super resolution},
\newblock in: \bibinfo{booktitle}{Proceedings of the IEEE International
  Conference on Computer Vision}, \bibinfo{year}{2017}, pp.
  \bibinfo{pages}{1689--1697}.
%Type = Article
\bibitem[{Goodfellow et~al.(2014)Goodfellow, Pouget-Abadie, Mirza, Xu,
  Warde-Farley, Ozair, Courville, and Bengio}]{goodfellow2014generative}
\bibinfo{author}{I.~J. Goodfellow}, \bibinfo{author}{J.~Pouget-Abadie},
  \bibinfo{author}{M.~Mirza}, \bibinfo{author}{B.~Xu},
  \bibinfo{author}{D.~Warde-Farley}, \bibinfo{author}{S.~Ozair},
  \bibinfo{author}{A.~Courville}, \bibinfo{author}{Y.~Bengio},
\newblock \bibinfo{title}{Generative adversarial networks},
\newblock \bibinfo{journal}{arXiv preprint arXiv:1406.2661}
  (\bibinfo{year}{2014}).
%Type = Inproceedings
\bibitem[{Indradi et~al.(2019)Indradi, Arifianto, and
  Ramadhani}]{indradi2019face}
\bibinfo{author}{S.~D. Indradi}, \bibinfo{author}{A.~Arifianto},
  \bibinfo{author}{K.~N. Ramadhani},
\newblock \bibinfo{title}{Face image super-resolution using inception residual
  network and gan framework},
\newblock in: \bibinfo{booktitle}{2019 7th International Conference on
  Information and Communication Technology (ICoICT)},
  \bibinfo{organization}{IEEE}, \bibinfo{year}{2019}, pp.
  \bibinfo{pages}{1--6}.
%Type = Inproceedings
\bibitem[{Yang et~al.(2020)Yang, Wang, Ma, Gao, Liu, Wang, and
  Ren}]{yang2020hifacegan}
\bibinfo{author}{L.~Yang}, \bibinfo{author}{S.~Wang}, \bibinfo{author}{S.~Ma},
  \bibinfo{author}{W.~Gao}, \bibinfo{author}{C.~Liu},
  \bibinfo{author}{P.~Wang}, \bibinfo{author}{P.~Ren},
\newblock \bibinfo{title}{Hifacegan: Face renovation via collaborative
  suppression and replenishment},
\newblock in: \bibinfo{booktitle}{Proceedings of the 28th ACM International
  Conference on Multimedia}, \bibinfo{year}{2020}, pp.
  \bibinfo{pages}{1551--1560}.
%Type = Article
\bibitem[{Jiang et~al.(2019)Jiang, Wang, Yi, Wang, Gu, and
  Jiang}]{jiang2019atmfn}
\bibinfo{author}{K.~Jiang}, \bibinfo{author}{Z.~Wang}, \bibinfo{author}{P.~Yi},
  \bibinfo{author}{G.~Wang}, \bibinfo{author}{K.~Gu},
  \bibinfo{author}{J.~Jiang},
\newblock \bibinfo{title}{Atmfn: Adaptive-threshold-based multi-model fusion
  network for compressed face hallucination},
\newblock \bibinfo{journal}{IEEE Transactions on Multimedia}
  \bibinfo{volume}{22} (\bibinfo{year}{2019}) \bibinfo{pages}{2734--2747}.
%Type = Article
\bibitem[{Chen et~al.(2020)Chen, Gong, Wang, Li, and Wong}]{chen2020learning}
\bibinfo{author}{C.~Chen}, \bibinfo{author}{D.~Gong},
  \bibinfo{author}{H.~Wang}, \bibinfo{author}{Z.~Li}, \bibinfo{author}{K.-Y.~K.
  Wong},
\newblock \bibinfo{title}{Learning spatial attention for face
  super-resolution},
\newblock \bibinfo{journal}{IEEE Transactions on Image Processing}
  \bibinfo{volume}{30} (\bibinfo{year}{2020}) \bibinfo{pages}{1219--1231}.
%Type = Article
\bibitem[{Mnih et~al.(2014)Mnih, Heess, Graves, and
  Kavukcuoglu}]{mnih2014recurrent}
\bibinfo{author}{V.~Mnih}, \bibinfo{author}{N.~Heess},
  \bibinfo{author}{A.~Graves}, \bibinfo{author}{K.~Kavukcuoglu},
\newblock \bibinfo{title}{Recurrent models of visual attention},
\newblock \bibinfo{journal}{arXiv preprint arXiv:1406.6247}
  (\bibinfo{year}{2014}).
%Type = Inproceedings
\bibitem[{Hu et~al.(2018)Hu, Shen, and Sun}]{hu2018squeeze}
\bibinfo{author}{J.~Hu}, \bibinfo{author}{L.~Shen}, \bibinfo{author}{G.~Sun},
\newblock \bibinfo{title}{Squeeze-and-excitation networks},
\newblock in: \bibinfo{booktitle}{Proc. of Comput. Vis. and Patt. Recogn.},
  \bibinfo{year}{2018}, pp. \bibinfo{pages}{7132--7141}.
%Type = Inproceedings
\bibitem[{Xu et~al.(2015)Xu, Ba, Kiros, Cho, Courville, Salakhudinov, Zemel,
  and Bengio}]{xu2015show}
\bibinfo{author}{K.~Xu}, \bibinfo{author}{J.~Ba}, \bibinfo{author}{R.~Kiros},
  \bibinfo{author}{K.~Cho}, \bibinfo{author}{A.~Courville},
  \bibinfo{author}{R.~Salakhudinov}, \bibinfo{author}{R.~Zemel},
  \bibinfo{author}{Y.~Bengio},
\newblock \bibinfo{title}{Show, attend and tell: Neural image caption
  generation with visual attention},
\newblock in: \bibinfo{booktitle}{International conference on machine
  learning}, \bibinfo{organization}{PMLR}, \bibinfo{year}{2015}, pp.
  \bibinfo{pages}{2048--2057}.
%Type = Inproceedings
\bibitem[{Wang et~al.(2017)Wang, Jiang, Qian, Yang, Li, Zhang, Wang, and
  Tang}]{wang2017residual}
\bibinfo{author}{F.~Wang}, \bibinfo{author}{M.~Jiang},
  \bibinfo{author}{C.~Qian}, \bibinfo{author}{S.~Yang},
  \bibinfo{author}{C.~Li}, \bibinfo{author}{H.~Zhang},
  \bibinfo{author}{X.~Wang}, \bibinfo{author}{X.~Tang},
\newblock \bibinfo{title}{Residual attention network for image classification},
\newblock in: \bibinfo{booktitle}{Proceedings of the IEEE conference on
  computer vision and pattern recognition}, \bibinfo{year}{2017}, pp.
  \bibinfo{pages}{3156--3164}.
%Type = Inproceedings
\bibitem[{Woo et~al.(2018)Woo, Park, Lee, and Kweon}]{woo2018cbam}
\bibinfo{author}{S.~Woo}, \bibinfo{author}{J.~Park}, \bibinfo{author}{J.-Y.
  Lee}, \bibinfo{author}{I.~S. Kweon},
\newblock \bibinfo{title}{Cbam: Convolutional block attention module},
\newblock in: \bibinfo{booktitle}{Proceedings of the European conference on
  computer vision (ECCV)}, \bibinfo{year}{2018}, pp. \bibinfo{pages}{3--19}.
%Type = Inproceedings
\bibitem[{Fu et~al.(2017)Fu, Zheng, and Mei}]{fu2017look}
\bibinfo{author}{J.~Fu}, \bibinfo{author}{H.~Zheng}, \bibinfo{author}{T.~Mei},
\newblock \bibinfo{title}{Look closer to see better: Recurrent attention
  convolutional neural network for fine-grained image recognition},
\newblock in: \bibinfo{booktitle}{Proceedings of the IEEE conference on
  computer vision and pattern recognition}, \bibinfo{year}{2017}, pp.
  \bibinfo{pages}{4438--4446}.
%Type = Inproceedings
\bibitem[{Zhang et~al.(2018)Zhang, Tian, Kong, Zhong, and
  Fu}]{zhang2018residual}
\bibinfo{author}{Y.~Zhang}, \bibinfo{author}{Y.~Tian},
  \bibinfo{author}{Y.~Kong}, \bibinfo{author}{B.~Zhong},
  \bibinfo{author}{Y.~Fu},
\newblock \bibinfo{title}{Residual dense network for image super-resolution},
\newblock in: \bibinfo{booktitle}{Proc. of Comput. Vis. and Patt. Recogn.},
  \bibinfo{year}{2018}.
%Type = Inproceedings
\bibitem[{Szegedy et~al.(2017)Szegedy, Ioffe, Vanhoucke, and
  Alemi}]{szegedy2017inception}
\bibinfo{author}{C.~Szegedy}, \bibinfo{author}{S.~Ioffe},
  \bibinfo{author}{V.~Vanhoucke}, \bibinfo{author}{A.~Alemi},
\newblock \bibinfo{title}{Inception-v4, inception-resnet and the impact of
  residual connections on learning},
\newblock in: \bibinfo{booktitle}{Proc. of AAAI Conf. Artifi. Intelli.},
  volume~\bibinfo{volume}{31}, \bibinfo{year}{2017}.
%Type = Inproceedings
\bibitem[{Lim et~al.(2017)Lim, Son, Kim, Nah, and Mu~Lee}]{lim2017enhanced}
\bibinfo{author}{B.~Lim}, \bibinfo{author}{S.~Son}, \bibinfo{author}{H.~Kim},
  \bibinfo{author}{S.~Nah}, \bibinfo{author}{K.~Mu~Lee},
\newblock \bibinfo{title}{Enhanced deep residual networks for single image
  super-resolution},
\newblock in: \bibinfo{booktitle}{Proc. of Comput. Vis. and Patt. Recogn.
  Worksh.}, \bibinfo{year}{2017}, pp. \bibinfo{pages}{136--144}.
%Type = Article
\bibitem[{Simonyan and Zisserman(2014)}]{simonyan2014very}
\bibinfo{author}{K.~Simonyan}, \bibinfo{author}{A.~Zisserman},
\newblock \bibinfo{title}{Very deep convolutional networks for large-scale
  image recognition},
\newblock \bibinfo{journal}{arXiv preprint arXiv:1409.1556}
  (\bibinfo{year}{2014}).
%Type = Inproceedings
\bibitem[{Deng et~al.(2009)Deng, Dong, Socher, Li, Li, and
  Fei-Fei}]{deng2009imagenet}
\bibinfo{author}{J.~Deng}, \bibinfo{author}{W.~Dong},
  \bibinfo{author}{R.~Socher}, \bibinfo{author}{L.-J. Li},
  \bibinfo{author}{K.~Li}, \bibinfo{author}{L.~Fei-Fei},
\newblock \bibinfo{title}{Imagenet: A large-scale hierarchical image database},
\newblock in: \bibinfo{booktitle}{2009 IEEE conference on computer vision and
  pattern recognition}, \bibinfo{organization}{Ieee}, \bibinfo{year}{2009}, pp.
  \bibinfo{pages}{248--255}.
%Type = Article
\bibitem[{Salimans et~al.(2016)Salimans, Goodfellow, Zaremba, Cheung, Radford,
  and Chen}]{salimans2016improved}
\bibinfo{author}{T.~Salimans}, \bibinfo{author}{I.~Goodfellow},
  \bibinfo{author}{W.~Zaremba}, \bibinfo{author}{V.~Cheung},
  \bibinfo{author}{A.~Radford}, \bibinfo{author}{X.~Chen},
\newblock \bibinfo{title}{Improved techniques for training gans},
\newblock \bibinfo{journal}{arXiv preprint arXiv:1606.03498}
  (\bibinfo{year}{2016}).
%Type = Inproceedings
\bibitem[{Szegedy et~al.(2016)Szegedy, Vanhoucke, Ioffe, Shlens, and
  Wojna}]{szegedy2016rethinking}
\bibinfo{author}{C.~Szegedy}, \bibinfo{author}{V.~Vanhoucke},
  \bibinfo{author}{S.~Ioffe}, \bibinfo{author}{J.~Shlens},
  \bibinfo{author}{Z.~Wojna},
\newblock \bibinfo{title}{Rethinking the inception architecture for computer
  vision},
\newblock in: \bibinfo{booktitle}{Proceedings of the IEEE conference on
  computer vision and pattern recognition}, \bibinfo{year}{2016}, pp.
  \bibinfo{pages}{2818--2826}.
%Type = Article
\bibitem[{Zhang et~al.(2016)Zhang, Zhang, Li, and Qiao}]{zhang2016joint}
\bibinfo{author}{K.~Zhang}, \bibinfo{author}{Z.~Zhang},
  \bibinfo{author}{Z.~Li}, \bibinfo{author}{Y.~Qiao},
\newblock \bibinfo{title}{Joint face detection and alignment using multitask
  cascaded convolutional networks},
\newblock \bibinfo{journal}{IEEE Signal Processing Letters}
  \bibinfo{volume}{23} (\bibinfo{year}{2016}) \bibinfo{pages}{1499--1503}.
%Type = Article
\bibitem[{Zhang et~al.(2011)Zhang, Zhang, Mou, and Zhang}]{zhang2011fsim}
\bibinfo{author}{L.~Zhang}, \bibinfo{author}{L.~Zhang},
  \bibinfo{author}{X.~Mou}, \bibinfo{author}{D.~Zhang},
\newblock \bibinfo{title}{Fsim: A feature similarity index for image quality
  assessment},
\newblock \bibinfo{journal}{IEEE transactions on Image Processing}
  \bibinfo{volume}{20} (\bibinfo{year}{2011}) \bibinfo{pages}{2378--2386}.
%Type = Article
\bibitem[{Lanaras et~al.(2018)Lanaras, Bioucas-Dias, Galliani, Baltsavias, and
  Schindler}]{lanaras2018super}
\bibinfo{author}{C.~Lanaras}, \bibinfo{author}{J.~Bioucas-Dias},
  \bibinfo{author}{S.~Galliani}, \bibinfo{author}{E.~Baltsavias},
  \bibinfo{author}{K.~Schindler},
\newblock \bibinfo{title}{Super-resolution of sentinel-2 images: Learning a
  globally applicable deep neural network},
\newblock \bibinfo{journal}{ISPRS Journal of Photogrammetry and Remote Sensing}
  \bibinfo{volume}{146} (\bibinfo{year}{2018}) \bibinfo{pages}{305--319}.
%Type = Article
\bibitem[{Wang and Bovik(2002)}]{wang2002universal}
\bibinfo{author}{Z.~Wang}, \bibinfo{author}{A.~C. Bovik},
\newblock \bibinfo{title}{A universal image quality index},
\newblock \bibinfo{journal}{IEEE signal processing letters} \bibinfo{volume}{9}
  (\bibinfo{year}{2002}) \bibinfo{pages}{81--84}.
%Type = Article
\bibitem[{Mittal et~al.(2012)Mittal, Moorthy, and Bovik}]{mittal2012no}
\bibinfo{author}{A.~Mittal}, \bibinfo{author}{A.~K. Moorthy},
  \bibinfo{author}{A.~C. Bovik},
\newblock \bibinfo{title}{No-reference image quality assessment in the spatial
  domain},
\newblock \bibinfo{journal}{IEEE Transactions on image processing}
  \bibinfo{volume}{21} (\bibinfo{year}{2012}) \bibinfo{pages}{4695--4708}.
%Type = Inproceedings
\bibitem[{Liu et~al.(2015)Liu, Luo, Wang, and Tang}]{liu2015faceattributes}
\bibinfo{author}{Z.~Liu}, \bibinfo{author}{P.~Luo}, \bibinfo{author}{X.~Wang},
  \bibinfo{author}{X.~Tang},
\newblock \bibinfo{title}{Deep learning face attributes in the wild},
\newblock in: \bibinfo{booktitle}{Proceedings of International Conference on
  Computer Vision (ICCV)}, \bibinfo{year}{2015}.
%Type = Inproceedings
\bibitem[{Lee et~al.(2018)Lee, Zhang, Lee, Cheng, and Hsu}]{lee2018attribute}
\bibinfo{author}{C.-H. Lee}, \bibinfo{author}{K.~Zhang}, \bibinfo{author}{H.-C.
  Lee}, \bibinfo{author}{C.-W. Cheng}, \bibinfo{author}{W.~Hsu},
\newblock \bibinfo{title}{Attribute augmented convolutional neural network for
  face hallucination},
\newblock in: \bibinfo{booktitle}{Proceedings of the IEEE conference on
  computer vision and pattern recognition workshops}, \bibinfo{year}{2018}, pp.
  \bibinfo{pages}{721--729}.

\end{thebibliography}





% \end{thebibliography}
\end{document}